%% file: acl_latex.tex
\pdfoutput=1
\documentclass[11pt]{article}

\usepackage[preprint]{acl}

\usepackage{times}
\usepackage{latexsym}

\usepackage[T1]{fontenc}
\usepackage[utf8]{inputenc}

\usepackage{microtype}

\usepackage{inconsolata}

\usepackage{graphicx}

\usepackage{enumitem}

\usepackage{fontawesome}

\usepackage[breakable,skins,theorems]{tcolorbox}
\usepackage{listings}
\tcbset{
  takeaway_style/.style={
    width=\linewidth,
    top=6pt,
    bottom=4pt,
    colback=white,
    colframe=blue!25,
    colbacktitle=blue!60!white!80,
    enhanced,
    center,
    attach boxed title to top left={yshift=-0.1in,xshift=0in},
    boxed title style={boxrule=0pt,colframe=white,},
  }
}
\newtcolorbox[auto counter,number within=subsection,]{takeaway}[2][]{
    takeaway_style,
    title={Insight \thetcbcounter},
    #1
}

\title{Does Context Matter? \textsc{ContextualJudgeBench} for Evaluating LLM-based Judges in Contextual Settings}

\author{Austin Xu$^\star$,  Srijan Bansal$^\star$,  Yifei Ming,  Semih Yavuz,  Shafiq Joty\\
Salesforce AI Research\\
{\small $^\star$ Co-lead, equal contribution. Correspondence: \{austin.xu, srijanbansal\}@salesforce.com}\\
\\
{\normalsize \faGithub~ \url{https://github.com/SalesforceAIResearch/ContextualJudgeBench}}\\
{\normalsize \includegraphics[height=11pt]{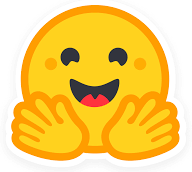}~ \url{https://huggingface.co/datasets/Salesforce/ContextualJudgeBench}}\\
\\
\\}

\newtoggle{comments}
\togglefalse{comments}

\iftoggle{comments}{
    \newcommand{\semih}[1]{\textcolor{blue}{(Semih: #1)}}
    \newcommand{\austin}[1]{\textcolor{orange}{(Austin: #1)}}
    \newcommand{\srijan}[1]{\textcolor{violet}{(Srijan: #1)}}
    \newcommand{\shafiq}[1]{\textcolor{cyan}{(shafiq: #1)}}
    \newcommand{\yf}[1]{\textcolor{red}{(yifei: #1)}}
}{
    \newcommand{\semih}[1]{}
    \newcommand{\austin}[1]{}
    \newcommand{\srijan}[1]{}
    \newcommand{\shafiq}[1]{}
    \newcommand{\yf}[1]{}
}

\newcommand{\cmark}{\ding{51}}%
\newcommand{\xmark}{\ding{55}}%
\newcommand{\name}{ContextualJudgeBench}

\usepackage{cleveref}
\crefname{section}{sec.}{sec.}%
\crefname{figure}{fig.}{figs.}%
\crefname{appendix}{app.}{app.}%
\crefname{table}{tab.}{tab.}%

\usepackage{booktabs}
\usepackage{multirow}
\usepackage{multicol}
\usepackage{makecell}
\usepackage{pifont}

\usepackage[T1]{fontenc}

\begin{document}
\maketitle

\input{src/00Abstract}
\input{src/01Introduction}

\input{src/02Background}
\input{src/03Benchmark}
\input{src/04Result}

\input{src/05Analysis}
\input{src/06Conclusion}

\section*{Limitations}
Our evaluations center around generative evaluators, as they are the most flexible in terms of incorporating context and indicating different evaluation criteria. However, reward models (RMs) are a common class of evaluators that may be applicable to this setting. However, to our knowledge, no contextual reward models exist. While in practice, one can embed the context in the input, it is unclear how to derive criteria-specific rewards from current models. A fruitful direction of future work is developing and benchmarking classifier based RMs for contextual settings.

As we repurposed existing annotated datasets -- particularly for faithfulness and completeness -- we are constrained by their coverage. This limitation may prevent us from making observations that generalize beyond their original distribution. Furthermore, \name{} is constructed primarily from English sources, a language abundant with context, model responses, and corresponding annotations. Further research should aim to rigorously assess contextual assessment in low-resource languages, where contextual content and corresponding annotations may be more scarce. 

% lacks human verification on pair 
% can be extended to more difficult exmaples

%\section*{Acknowledgments}

\bibliography{custom,anthology}

\clearpage
\newpage
\appendix
\input{src/10Appendix}

\end{document}

%% file: src/00Abstract.tex
\begin{abstract}
The large language model (LLM)-as-judge paradigm has been used to meet the demand for a cheap, reliable, and fast evaluation of model outputs during AI system development and post-deployment monitoring. While judge models---LLMs finetuned to specialize in assessing and critiquing model outputs---have been touted as general purpose evaluators, they are typically evaluated only on \textit{non-contextual scenarios}, such as instruction following. The omission of contextual settings---those where external information is used as \textit{context} to generate an output---is surprising given the increasing prevalence of retrieval-augmented generation (RAG) and summarization use cases. Contextual assessment is uniquely challenging, as evaluation often depends on practitioner priorities, leading to conditional evaluation criteria (e.g., comparing responses based on factuality and then considering completeness if they are equally factual).
To address the gap, we propose~\name{},
a judge benchmark with 2,000 challenging response pairs across eight splits inspired by real-world contextual evaluation scenarios. We build our benchmark with a multi-pronged data construction pipeline that leverages both existing human annotations and model-based perturbations. Our comprehensive study across 11 judge models and 9 general purpose models, reveals that the contextual information and its assessment criteria present a significant challenge to even state-of-the-art models. For example, OpenAI’s o1, the best-performing model, barely reaches 55\% consistent accuracy.
\end{abstract}

%% file: src/01Introduction.tex
\section{Introduction}\label{sec:intro}
In the LLM era, timely, affordable, and accurate evaluation of model responses is essential for model development and monitoring. One automated evaluation solution available to practitioners is the \textit{LLM-as-judge} approach, where relatively lightweight \textit{judge models} are trained to evaluate and critique other model responses. Judge models are broadly touted as general-purpose evaluators (e.g., \citet{vu2024foundational,alexandru2025atla}), capable of being deployed across domains and evaluation settings. However, judges are rarely evaluated on \textit{contextual settings} \citep{wang2024self,saha2025learning,ye2024beyond}, where the evaluated responses are generated from an externally provided context rather than solely from the model's parametric knowledge, like in retrieval-augmented generation (RAG) or summarization.

\begin{figure}[t]
    \centering
    \includegraphics[width=0.9\columnwidth]{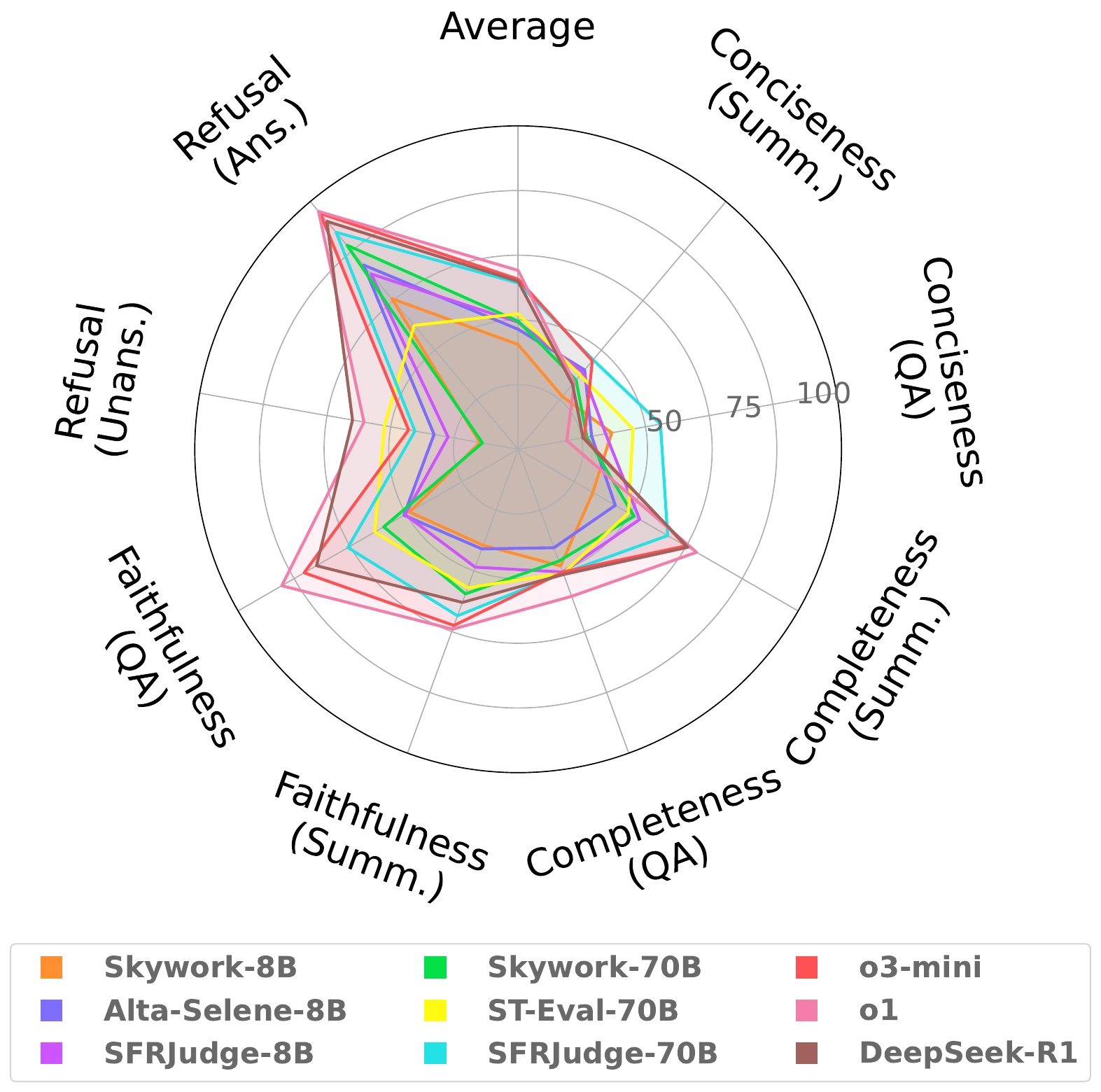}
   
    \vspace{0.5\baselineskip}
    \includegraphics[width=0.9\columnwidth]{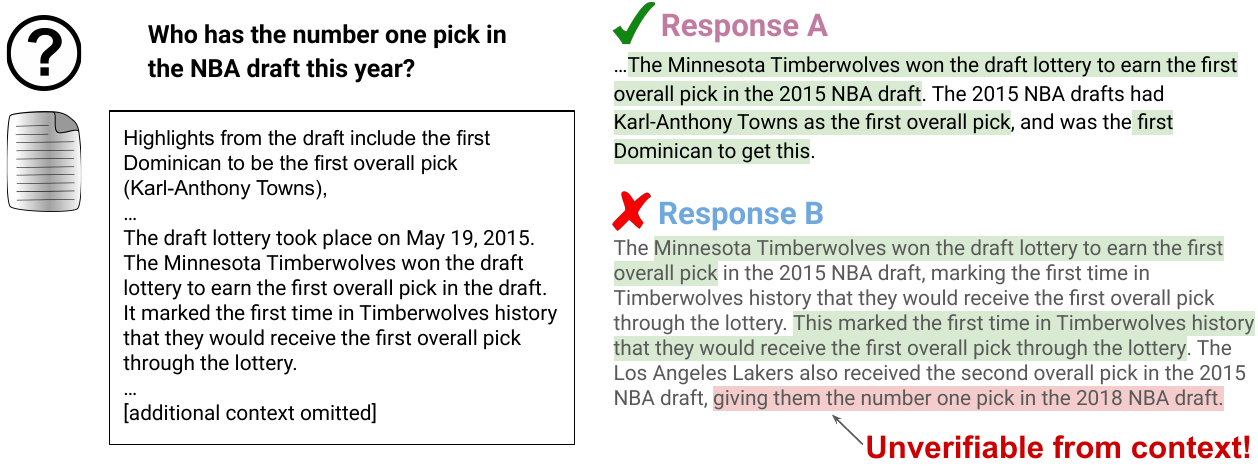}
    \caption{(Top) An overview of top-performing models on the eight splits of ContextualJudgeBench. (Bottom) A truncated sample from the faithfulness split, where Response A is preferred because all of its content is factually verifiable from the context. 
    }
     \label{fig:spider_plot}
\end{figure}

As contextual generation systems gain prominence, specialized generators~\citep{commandr,rag2,nguyen2024sfr} have been developed to meet the stringent faithfulness demands of business applications and high-risk fields, like medicine~\citep{xiong2024benchmarking} and law~\citep{wiratunga2024cbr}. Reliably evaluating such systems is increasingly important, but presents unique challenges. The presence of contextual information magnifies challenges that exist in non-contextual human evaluation~\citep{liu2023benchmarking}: Since contextual generation requires responses to be \textit{faithful} to the provided context, humans must first comprehend potentially long, domain-specific contexts before they can evaluate a response. This additional ``hallucination detection'' step adds another layer of complexity on top of evaluating the substantive quality of responses.

Taken together, contextual settings are the ideal candidate for automatic evaluation: LLMs have strong language understanding across specialized domains~\cite{xie2023efficient,ke2025demystifying,colombo2024saullm} and have rapidly improving long-context comprehension abilities~\cite{kamradt2023pressure}. Indeed, many recent benchmarks for contextual generation use prompted~\citep{laban2024summary,jacovi2025facts} or finetuned~\citep{friel2024ragbench} LLMs to serve as evaluators due to longer, more complex model outputs. 
However, to our knowledge, no benchmarks exist to measure the quality of \textit{contextual evaluators.} We bridge this gap by proposing \name{}, which consists of 2,000 challenging pairwise samples across 8 splits that measure different evaluation criteria and settings.~\Cref{fig:spider_plot} showcases our dataset splits and benchmarking results. Our work \textit{complements} existing contextual generation benchmarks by offering a way to assess contextual evaluators.  

The dominant criteria for the contextual evaluation of responses center around \textit{faithfulness} and \textit{answer relevancy}~\citep{es2023ragas,saad2023ares,jacovi2025facts,laban2024summary}. Such metrics are often assigned independently in a {pointwise} manner, i.e., a model assigns a faithfulness score and a relevance score to a single response, with each score assigned without considering the other. \name{}, in contrast, 
proposes a pairwise evaluation setup. 
This pairwise setup offers utility to practitioners (e.g., evaluation for A/B testing) while eliciting evaluations better aligned with humans judgment from automatic evaluators~\citep{wang2023large,liu2024aligning}. However, directly using pointwise scores to do pairwise comparisons can lead to ambiguity: If a response is more relevant but less faithful, is it better? 

To remedy this, we propose a principled \textit{conditional} evaluation hierarchy (\Cref{sec:benchmark}) that prioritizes refusal accuracy and response faithfulness. First, we evaluate if judges can assess accurate or inaccurate refusals, where a response that refuses to answer due to a perceived lack of evidence is compared against a substantive response. Given two substantive responses, we next assess based on faithfulness: Which response contains more factually supported information? If two responses are equally faithful, then they are evaluated on completeness, with more thorough responses being preferred. Finally, for two equally complete responses, they are evaluated based on conciseness, as responses should not contain extraneous information, even if factual. The splits in \name{} are carefully designed to test judges in each setting that arises in this hierarchy. Concretely, our contributions are:
\begin{itemize}[leftmargin=*,noitemsep,topsep=5pt]
    \item With an emphasis on refusals and faithfulness, we propose a hierarchy that provides an ``order of operations'' for pairwise contextual evaluation.
    \item We present \name{}, a benchmark for evaluating judge models consisting of 2,000 response pairs across eight splits derived from real-world contextual outcomes.
    \item We evaluate 11 judge models, ranging in size from 3.8B to 70B parameters on \name{} along with 9 general purpose/reasoning models. 
\end{itemize}
 Our findings reveal that contextual assessment remains an open challenge, with GPT-o1 and SFRJudge-70B \citep{wang2024direct} only achieving 55.3 and 51.4 accuracy. Despite the reasoning intensive nature of contextual evaluation, our analysis shows that inference-time scaling for judges may actually lead to performance \textit{degradations}. \austin{More interesting comments?}

%% file: src/02Background.tex
\section{Related work}
Our work, rather than evaluating contextual systems, evaluates judge models as contextual \textit{evaluators}. Here, we review current judge benchmarks and contextual evaluation setups.\\

\noindent\textbf{Evaluation for LLM-as-judges.}
LLM-as-judge is a generative evaluator paradigm where LLMs are trained to produce an evaluation (natural language explanation and judgment) given the original user input, evaluation protocol (rules and criteria for evaluation), and model responses as input. As the popularity of LLM-as-judges grows, numerous benchmarks have been proposed to evaluate these evaluators. These benchmarks are typically for specific domains, like instruction following~\citep{zeng2023evaluating}, fine-grained evaluation~\citep{kim2023prometheus,kim2024prometheus}, bias~\citep{park2024offsetbias}, reward modeling~\citep{lambert2024rewardbench,frick2024evaluate,gureja2024m}, or reasoning~\citep{tan2024judgebench}. While new judge benchmarks are challenging, none focus on contextual evaluation. Of judge benchmarks, a subset of Eval-P~\citep{li2023generative} contains summarization pairs with the winner chosen by aggregating various criteria into an overall score. InstruSum~\citep{liu2023benchmarking} has also been used for judge evaluation~\citep{wang2024direct,alexandru2025atla,liu2024reife}.
\name{}, in contrast, is dedicated entirely to contextual evaluation, requiring evaluation to be done in under-explored settings like RAG-QA along previously untested criteria such as refusal.\\

\noindent\textbf{Evaluation for contextual responses.}
RAG generators have been typically evaluated with standard knowledge-based QA tasks, e.g., ContextualBench~\citep{nguyen2024sfr}, or with newer benchmarks that cover scenarios such as faithfulness~\citep{{ming2024faitheval,niu2023ragtruth,tang2024minicheck,li2023halueval,sadat2023delucionqa}}, diverse domains~\citep{friel2024ragbench}, refusals~\citep{peng2024unanswerability}, and reasoning~\citep{wei2024measuring,krishna2024fact}. Because RAG settings have progressed beyond simple factoid answers, recent benchmarks have deployed carefully prompted frontier LLMs (e.g.,~\citet{jacovi2025facts}) to perform assessment in a pointwise manner, rather than using exact string matching~\citep{nguyen2024sfr}.

Initial evaluation efforts for RAG settings focused on faithfulness, training hallucination detectors~\citep{tang2024minicheck} as both sequence classifiers and generative models~\citep{wang2024halu,ravi2024lynx,ramamurthy2024veritas}. More holistic evaluation systems with multiple metrics have recently been proposed, such as \citet{es2023ragas,saad2023ares}. For the most part, these approaches involve specialized prompting~\citep{es2023ragas}, using synthetic data generation to train specialized evaluators~\citep{saad2023ares}. 

Summarization evaluation has evolved from n-gram metrics like ROUGE~\citep{lin2004rouge} and METEOR~\citep{banerjee-lavie-2005-meteor} to contextual embedding model scorers~\citep{zhang2020bertscoreevaluatingtextgeneration, zhao2019moverscore, yuan2021bartscoreevaluatinggeneratedtext}. 
However, these evaluators cannot assess based on multiple criteria and tend to correlate poorly with humans. 
To evaluate quality, the primary focus has been on model-based factual verification~\citep{laban-etal-2022-summac, cao-wang-2021-cliff, goyal-durrett-2021-annotating, kryscinski-etal-2020-evaluating, laban-etal-2023-summedits}. 
Recent studies have shifted toward human annotations for finer-grained assessment~\citep{song-etal-2024-finesure, lee-etal-2024-unisumeval, oh-etal-2025-learning}, focusing on metrics such as faithfulness and conciseness. As summarization has become more instruction-controllable, LLM evaluators have been tasked with more controlled assessments~\citep{liu2023benchmarking, laban2024summary}.

Our proposed work complements these existing benchmarks in summarization and RAG by evaluating contextual \textit{judges}, rather than the generators.

%% file: src/03Benchmark.tex
\section{\name{}}\label{sec:benchmark}
\begin{figure*}[t]
    \centering
    \includegraphics[width=0.85\linewidth]{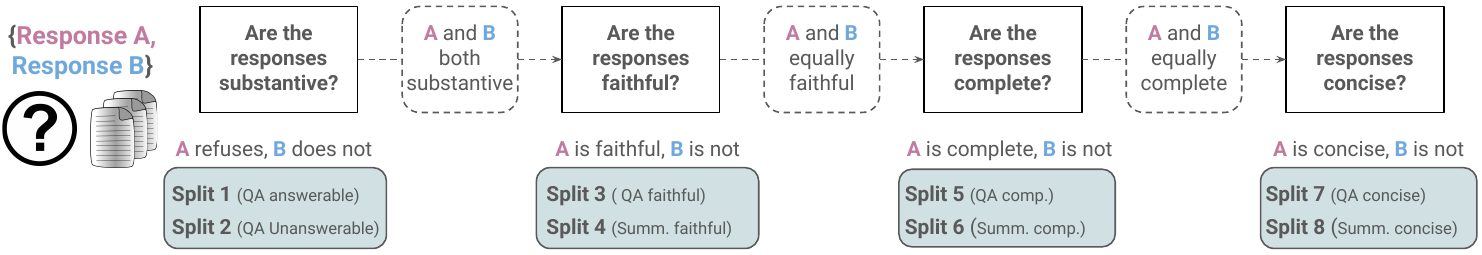}
    \caption{A refusal and faithfulness-first contextual evaluation hierarchy, as assessed by ContextualJudgeBench. 
    } 
    \label{fig:workflow}
    
\end{figure*}

\begin{quote}
    \textit{``63 percent of respondents...said that output inaccuracy was the greatest risk they saw in their organizations’ use of gen AI.''}
    --\href{https://www.mckinsey.com/capabilities/mckinsey-digital/our-insights/a-data-leaders-technical-guide-to-scaling-gen-ai}{McKinsey, 2024 AI Survey}
\end{quote}
Inaccuracy is the largest reported risk for practitioners using AI systems. 30\% of respondents in a \href{https://www2.deloitte.com/content/dam/Deloitte/us/Documents/consulting/us-state-of-gen-ai-q4.pdf}{Deloitte survey} specifically cite trust loss due to hallucinations as a top concern. Hallucinations are especially unacceptable in contextual settings, as the model is expected to generate responses strictly based on the provided context. This grounding context is typically considered a gold-standard source of knowledge. If the relevant information is absent, the model should refrain from responding rather than generate unsupported content.
Motivated by real-world concerns, we propose a conditional evaluation workflow
(\Cref{fig:workflow}) that prioritizes \textit{answerability} and \textit{faithfulness} before assessing other criteria. Each evaluation step in our workflow requires creating new splits for \name{}.

In developing contextual systems, practitioners often conduct A/B testing between systems with different generator, retriever,  pre-processing configurations~\citep{saad2023ares}. \name{} is designed to reflect this pairwise A/B testing setup, containing 2,000 test samples. Each sample includes a user input, a context, and two responses, from which a judge selects the ``better'' response based on our workflow.
The pairwise setting is well-suited for judge-based evaluation as it aligns closely with human preferences~\citep{wang2023large,liu2024aligning}. 
We first describe two methods we use to create the pairwise samples. Then, we present \name{} in four stages (\Cref{step1} -- \ref{step4}), each corresponding to a step in the evaluation workflow (\Cref{fig:workflow}). 

\subsection{Dataset creation approach}\label{sec:dataset-creation}

We employ two primary approaches to create \name{}: utilizing existing human annotations and leveraging frontier models for criteria-based response perturbation.
\begin{itemize}[leftmargin=*,noitemsep,topsep=5pt]
    \item \textbf{Human annotations [H]}:  We use existing human annotations \citep{lee-etal-2024-unisumeval, wan2024positionalbiasfaithfulnesslongform,wu2023finegrainedhumanfeedbackgives,liu-etal-2024-benchmarking} that evaluate multiple model responses for the same context. These assessments include criteria-specific scores or errors, either holistically or sentence-level. We select responses with significant differences based on specific criterion to form pairs, enabling comparative assessments.
    
    \item \textbf{Model-based perturbations}: In the absence of human labels, we form pairs through criteria-based response perturbation. Specifically, we use frontier LLMs to modify accurate responses based on the context to produce responses that do not align with the intended criteria. We apply this approach in two distinct ways:
    
    \textbf{Desired output prompting [M1]}: We ask an LLM to directly generate a response based on the context that fits certain output criteria. This includes generating context-based refusals or deliberately unfaithful responses.
    
    \textbf{Existing output modification [M2]}: We use an LLM to modify an existing response, introducing deviations based on predefined criteria. This can include making the response more verbose or altering its content in specific ways.
\end{itemize}

See \Cref{app:add_details} for details on the \name{} datasets, including data sources, pair sampling approaches, prompts used, and representative examples for each split.

\subsection{Step 1: QA refusal validity [splits 1 \& 2]} \label{step1}
Knowing when to refuse to answer due to lack of information is a critical first step specific to RAG settings.\footnote{
Refusals are uncommon in summarization, as instructions and context are both user provided; In RAG settings, the user has no control over the retrieved context.} 
Refusals can be viewed as a form of faithfulness: To remain faithful to the context, the model should refuse to hallucinate an answer if no relevant information is present. Conversely, the model should not refuse if the context is sufficient. 

Splits 1 and 2 of \name{} assess if judges can identify appropriate refusals. Each sample consists of a refusal (e.g., ``The answer cannot be answered based on the context'') and a substantive response. Split 1 contains answerable questions from LFRQA \cite{han2024rag}, where the judge should pick the substantive response, whereas split 2 contains unanswerable questions from FaithEval~\citep{ming2024faitheval}, making refusal the correct choice. To construct split 1,  we use approach \textbf{M1} from~\Cref{sec:dataset-creation}, using an LLM to generate context-based refusals as negative responses
to pair up with the provided positive responses. In split 2, we again employ approach \textbf{M1} to generate context-based refusal responses to correctly decline the question as positive responses and generate hallucinated (incorrect) responses as negative ones. See \Cref{prompt:p1} for generation prompt.

\subsection{Step 2: Faithfulness [splits 3 \& 4]}\label{step2}
When evaluating two substantive responses, the first criterion is \textit{faithfulness}, as a response cannot be considered accurate if it contains hallucinated content. Faithfulness measures the consistency of the response with the context: all factual statements in a faithful response must be attributable to the context, ensuring there are no hallucinations. Splits 3 and 4 evaluate the judge's ability to select the more faithful response for QA and summarization, respectively. Each pairwise sample is designed to include one substantively more faithful response, allowing the judge to choose the better response based solely on faithfulness. 

We construct split 3 by combining multiple QA datasets. For QA-Feedback \citep{wu2023finegrainedhumanfeedbackgives} and RAGTruth \citep{niu2023ragtruth}, we use the approach \textbf{H} to form pairs between RAG responses, annotated with either faithfulness scores or factuality errors. For LFRQA \citep{han2024rag}, LFQA \citep{xu-etal-2023-critical}, and short queries from MRQA \citep{fisch-etal-2019-mrqa}, we treat the provided responses as factually correct (positive) and apply approach \textbf{M1} to generate factually inconsistent negative responses based on the context. See \Cref{prompt:p1} for prompt template. We manually reviewed the formed pairs to ensure their reliability. 
%\shafiq{yes, mention some verification to give trust}\srijan{Done}. 
For Split 4, we use approach \textbf{H} to create summarization response pairs of different factuality levels. To ensure diversity, we sample contexts from \citet{wan2024positionalbiasfaithfulnesslongform, lee-etal-2024-unisumeval}, which cover both topic-specific and general summarization instructions across diverse domains.

\begin{figure}[t]
    \centering
    \includegraphics[width=0.9\columnwidth]{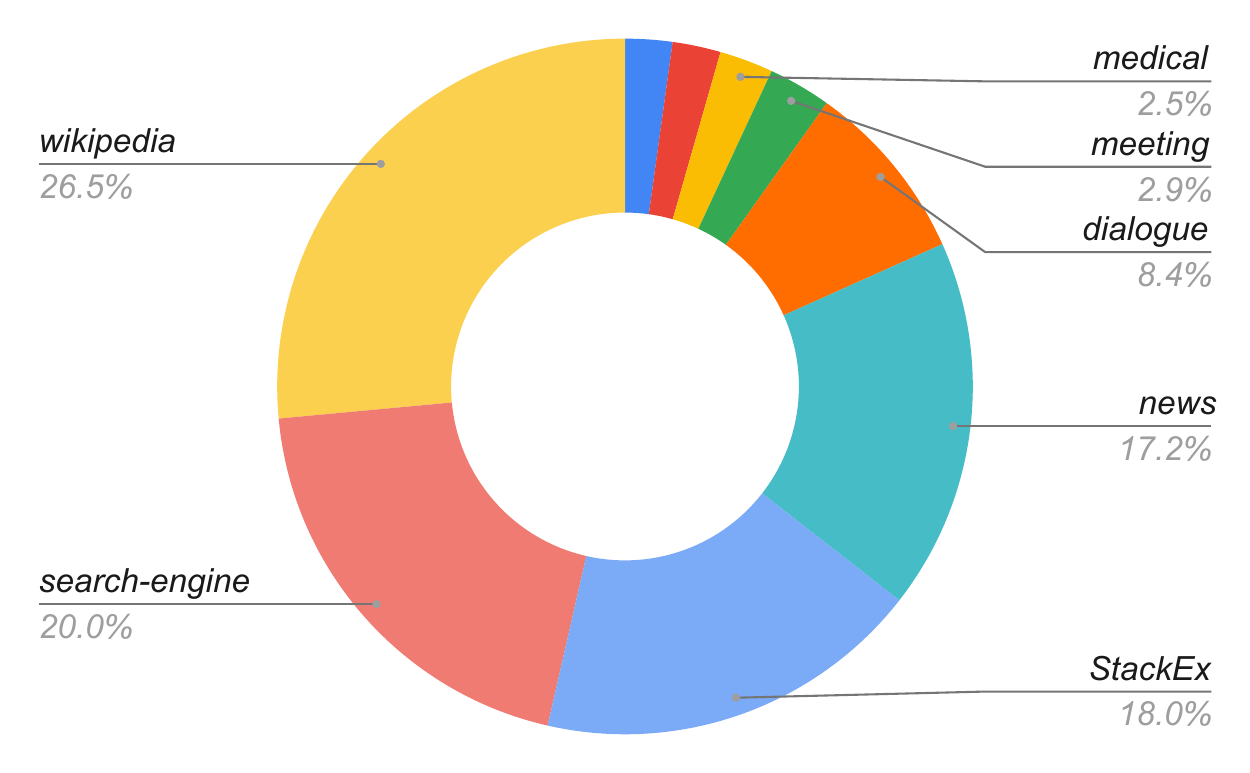}
    \caption{Distributions of context domain as a percent of the total
set of preference pairs in the benchmark.}
    \label{fig:domain_plot}
    
\end{figure}

\subsection{Step 3: Completeness [splits 5 \& 6]} \label{step3}
Beyond faithfulness, contextual evaluation must also assess response quality. When comparing two faithful responses, the better one should cover all essential information needed for a thorough and useful answer. As such, we consider \textit{completeness}, i.e., how comprehensive the response is, as the next criteria. Splits 5 and 6 assess the judge's ability to select more complete response when both options are faithful, for QA and summarization tasks, respectively. Each pairwise sample is designed such that one response is more complete than the other while both the responses are faithful. 

Judges should first confirm that both responses are faithful and then determine which one is more complete. We construct Split 5 using the LFRQA \citep{han2024rag} and QA-Feedback \citep{wu2023finegrainedhumanfeedbackgives} datasets. For LFRQA, we use approach \textbf{M2} from~\Cref{sec:dataset-creation} to modify a faithful response by omitting lines associated with certain citations while expanding on other citations. This yields a less complete negative response that is still faithful and similar in length to the original (positive) response. See \Cref{prompt:p1} for generation prompt. 
For QA-Feedback, we use approach \textbf{H} to create preference pairs from RAG responses annotated for completeness scores or missing information errors. Similarly, split 6 is created using approach \textbf{H} with existing human annotations that assess faithfulness and completeness in summarization responses. To form preference pairs, we first filter unfaithful responses. 
Then, we form pairs based on completeness, ensuring that one response is significantly more complete (positive) than the other (negative).

\subsection{Step 4: Conciseness [splits 7 \& 8]} \label{step4}

Our final criterion is \textit{conciseness}: does the response avoid including more than what was asked? Our hierarchy intentionally places conciseness after completeness, as an answer should not sacrifice relevant content for the sake of brevity. However, complete responses may not be \textit{minimally} complete: They may contain faithful yet extraneous information, repeated content, or unnecessary stylistic details.
In splits 7 and 8, each pairwise sample has one response that is more concise while maintaining the same faithfulness and completeness. Judges should first verify both responses are faithful and complete, then choose the more concise one.

% \semih{Is not Split-7 for QA based on the prev paragraph?}\srijan{fixed}
For Split 7, we use LFRQA~\citep{han2024rag} and QA-Feedback~\citep{wu2023finegrainedhumanfeedbackgives}. For LFRQA, we apply approach \textbf{M2}, tasking the model to insert direct quotations from the context without modifying the substance of provided responses. See~\Cref{prompt:p1} for generation prompt. For QA-Feedback, we use approach \textbf{H} to create pairs from responses annotated along conciseness, redundancy, and irrelevance. Preference pairs are formed by pairing faithful and complete responses by conciseness. For split 8, we again use approach \textbf{H}, using human annotations \citep{lee-etal-2024-unisumeval, liu-etal-2024-benchmarking} that assess summarization faithfulness, completeness, and conciseness.

%%%%

\subsection{Overall dataset statistics}

\name{} is constructed based on our evaluation workflow (Fig. \ref{fig:workflow}), resulting in 8 splits across 4 evaluation criteria, covering two common use cases of contextual generation: RAG-QA (5 splits) and Summarization (3 splits). We present the domain distribution in~\Cref{fig:domain_plot} and dataset statistics in~\Cref{tab:dataset_splits}. Overall, \name{} consists of 2,000 preference pairs, balanced across all splits, with over 1,500 unique contexts to minimize duplication. We include a wide range of context lengths, from a few tokens to nearly 10K tokens, with summarization contexts typically longer than QA ones. Response lengths range from brief answers to summaries over 1,000 tokens. To account for length bias in judges~\citep{zeng2023evaluating,park2024offsetbias},
we ensure minimal length differences between positive and negative responses across all splits; however, conciseness correlates with response length, resulting in longer positive responses.

\begin{table}[t]
    \centering
    \resizebox{0.996\columnwidth}{!}{
    \begin{tabular}{lcccccc} \toprule        
        \textbf{Split} & \textbf{\# Pairs} & \textbf{\# Context} & $\mathbf{{L}_{c}}$ & $ \mathbf{{L}_{r}}$ & $\mathbf{{L}_{pos}}$ & $\mathbf{{L}_{neg}}$  \\
        \midrule
        Refusal (Ans.) & 250 & 250 & 1,444 & 102 & 108 & 95 \\
        Refusal (Unans.) & 250 & 250 & 418 & 64 & 64 & 63 \\
        Faithfulness (QA) & 250 & 213 & 414 & 100 & 99 & 101 \\
        Faithfulness (Summ.) & 250 & 192 & 1,754 & 94 & 97 & 91 \\
        Completeness (QA) & 250 & 250 & 658 & 106 & 98 & 113 \\
        Completeness (Summ.) & 251 & 171 & 1,066 & 91 & 93 & 89 \\
        Conciseness (QA) & 255 & 254 & 1,086 & 199 & 116 & 281 \\
        Conciseness (Summ.) & 244 & 117 & 1,557 & 98 & 77 & 118 \\  
        \midrule
       \textbf{Total} & \textbf{2,000} & \textbf{1,537} & \textbf{1,048} & \textbf{107} & \textbf{94} & \textbf{119}  \\ 
        \bottomrule
    \end{tabular}
    }
    \caption{\name{} statistics. \# Context denotes unique contexts across all pairs. % (\# Pairs). 
    $\mathbf{{L}_{c}}$ and $\mathbf{{L}_{r}}$ represent the mean context and response lengths, while $\mathbf{{L}_{pos}}$ and $\mathbf{{L}_{neg}}$ denote the mean positive and negative response lengths per split.}
    \label{tab:dataset_splits}
    
\end{table}

%% file: src/04Result.tex
\section{Evaluation and analysis}\label{sec:results}
\subsection{Evaluation setup and baselines}\label{sec:eval-setup}
Because the order of responses influences judge decisions~\citep{wang2023large}, we adopt a consistency evaluation setup, like~\citet{tan2024judgebench,li2023generative}. We run evaluation for each test sample twice, swapping the order of responses for the second run. We denote these as \textit{Run 1} and \textit{Run 2}, respectively. Given the judge outputs of Run 1 and Run 2, we compute the following metrics.
\begin{itemize}[leftmargin=*,noitemsep,topsep=5pt]
    \item \textbf{Consistent accuracy}: A judge output is considered correct if the judge selects the correct response for both runs. Under this setup, randomly choosing responses achieves a consistent accuracy of 25\%. \textit{Our main evaluation results report consistent accuracy}.
    \item \textbf{Run 1 and 2 accuracy}: A judge output is considered correct is the judge selects the correct response in the respective run. These metrics do not take consistency into account, and may reflect more practical settings where inference can only be run once.
    \item \textbf{Optimistic accuracy}: A judge output is considered correct if \textit{either} the Run 1 or Run 2 output is correct, regardless of if the judge is consistent. This provides an optimistic upper bound for judge performance.
    \item \textbf{Consistency}: Consistency is the fraction of times a judge selects the same response in both runs, regardless of correctness.
\end{itemize}
\semih{(Maybe Future): Inconsistent judgment  might be a good qualitative example candidate, especially with judges providing explanation, showing how the explanations conflict, etc.}
To support a systemic investigation into positional bias (\Cref{sec:pos-bias-intro}), Run 1 sets Response A as the positive response  while Run 2 sets Response B as the positive response for all samples.

\begin{table}[t]
    \centering
    \resizebox{0.996\columnwidth}{!}{
    \begin{tabular}{lccc}\toprule
        \textbf{Model} & \textbf{\# Params} & \textbf{Expl.} & \textbf{Context len.} \\
        \midrule
        GLIDER \citep{deshpande2024glider} & 3.8B & \cmark & 128K \\
        Prometheus-2 \citep{kim2024prometheus} & 7,8x7B &  \cmark & 16K \\
        OffsetBias \citep{park2024offsetbias} & 8B & \xmark & 8K \\
        Atla-Selene \citep{alexandru2025atla} & 8B & \cmark & 128K \\
        Skywork-Critic  \citep{skyworkcritic2024} & 8,70B & \xmark & 128K\\ 
        SFRJudge  \citep{wang2024direct} & 8,12,70B & \cmark & 128K\\
        STEval. \citep{wang2024self} & 70B & \cmark & 128K\\
        \midrule
        Llama-3.1  \citep{dubey2024llama} & 8,70B & \cmark & 128K \\
        Llama-3.3  \citep{dubey2024llama} & 70B & \cmark & 128K \\
        GPT-4o,4o-mini  \citep{hurst2024gpt} & ? & \cmark & 128K\\
        GPT-o1,o3-mini  \citep{jaech2024openai} & ? & \cmark & 128K \\
        DeepSeek-R1  \cite{guo2025deepseek} & 685B & \cmark & 128K \\
        DeepSeek-R1-distill  \cite{guo2025deepseek} & 70B & \cmark & 128K \\
        \bottomrule
    \end{tabular}
    }    
    \caption{Judge (top) and general (bottom) models evaluated. \texttt{Expl.} denotes if model outputs explanations.
    }
    \label{tab:judge-models}
\end{table}

\input{figures/main_results}

We evaluate 11 competitive LLM-as-judge models, ranging in size from 3.8B to 70B parameters: Prometheus~\citep{kim2024prometheus}, OffsetBias~\citep{park2024offsetbias}, SFRJudge~\citep{wang2024direct}, Skywork-Critic~\citep{skyworkcritic2024}, Self-taugh-evaluator~\citep{wang2024self}, GLIDER~\citep{deshpande2024glider}, and Atla-Selene~\citep{alexandru2025atla}. See~\Cref{tab:judge-models} for an overview of judges and~\Cref{app:judge-model-details}  for a more detailed description of each evaluated judge. For each judge, we retain the original prompt template %provided by the model developer 
while modifying evaluation instructions to align with our proposed workflow. Please see~\Cref{app:judge-prompts} for prompt samples. In addition to specialized judges, we use the instruct versions of Llama-3.1-8B and 70B and Llama-3.3-70B, along with GPT-4o, GPT-4o-mini, o3-mini, o1, Deepseek-R1, and DeepSeek-R1-Llama-Distill as prompted judge model baselines.
For all non-reasoning model-based judges, we generate with greedy sampling. 

As a reference point, we also run RAGAS~\citep{es2023ragas}, a pointwise RAG evaluator that leverages both prompted frontier models and embedding models, as well as MiniCheck~\citep{tang2024minicheck}, a hallucination detector. 
We apply these two methods to benchmark splits covered by their respective metrics: refusal and faithfulness for both, and completeness for RAGAS. 
For RAGAS, we score each response pointwise and derive corresponding pairwise outcomes in line with our hierarchy (e.g., for the completeness split, two responses must be considered equally faithful). 
For MiniCheck, we directly compare the classifier probabilities of each response to determine the pairwise winner.

\begin{figure*}[t]
    \centering
    \includegraphics[width=0.9\textwidth]{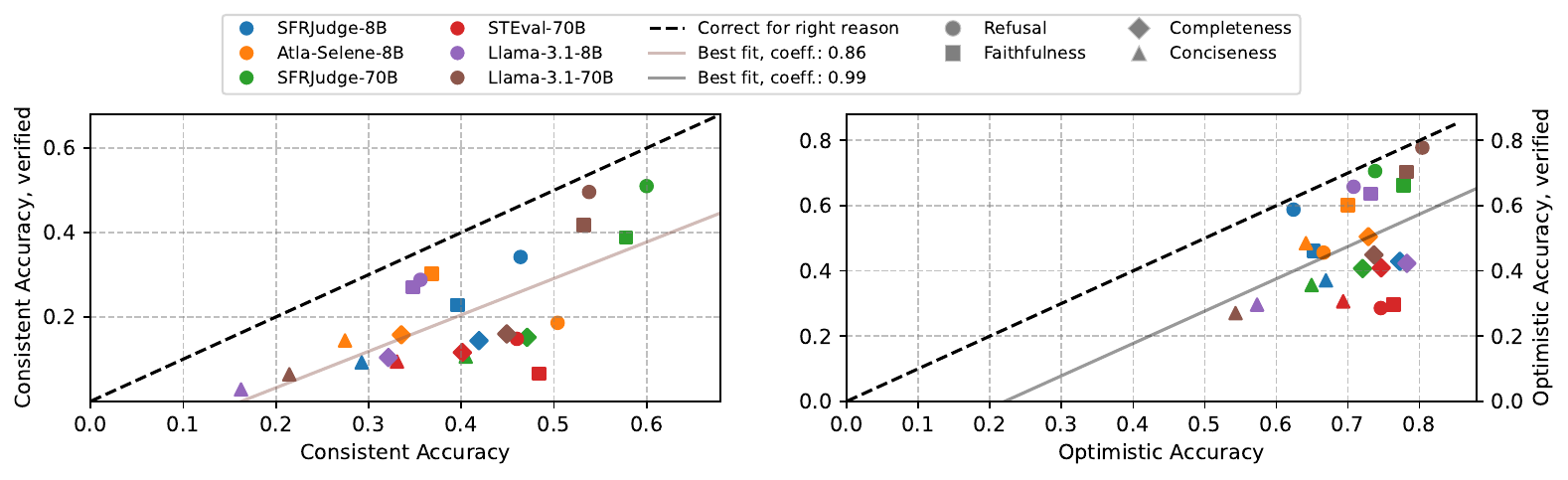}
    \caption{
    (Left) Accuracy vs. verified accuracy and (Right) optimistic accuracy vs. verified optimistic accuracy for six models, aggregated by criteria. The larger the drop from the dashed black line, the larger fraction of correct outcomes used incorrect criteria, as assessed by GPT-4o. 
    }
    \label{fig:critique_eval}
\end{figure*}

\begin{figure*}[t]
    \centering
    \includegraphics[width=0.9\textwidth]{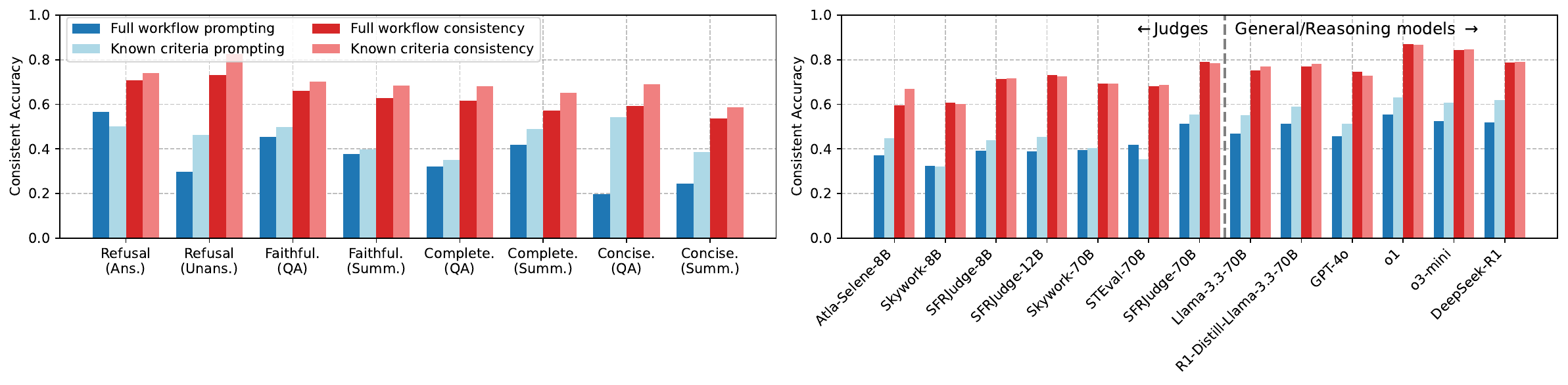}
    
    \caption{Judge performance changes are minor when given exact criteria vs. full workflow, indicating challenges in contextual evaluation beyond criteria. Per-split metrics (Left) averaged across all models, per-judge metrics (Right) averaged across all splits for a selected subset of judges.
    }
    \label{fig:criteria-specific}
\end{figure*}

\subsection{Judge model evaluation}\label{sec:main-results}
The results presented in~\Cref{tab:main-results} highlight the challenges of contextual evaluation. Overall, the best models on \name{} are o1 (55.3), o3-mini (52.6) and DeepSeek-R1 (51.9), two large-scale \textit{reasoning} models. The best-performing judge, SFRJudge-70B (51.4), nearly matches DeepSeek-R1. Judge model performance generally increases with model size, with the best-performing judges exceeding their similarly-sized API counterparts (e.g., SFRJudge-8B at 39.3 and GPT-4o-mini at 38.8). The scaling trend, along with the strong performance of reasoning models, suggests that contextual evaluation is a reasoning-intensive task. 

The performance difference between Llama-3.3-70B and its DeepSeek-R1 distilled counterpart clearly demonstrates the reasoning-intensive nature of contextual evaluation: DeepSeek-R1-Llama-70B outperforms its base model, Llama-3.3-70B-Instruct, by 4.4 points, with the only difference between the two models being reasoning-specific training. Despite the importance of strong reasoning ability, we show that two inference-time scaling techniques typically used in reasoning settings, self-consistency and better chain-of-thought prompting, do not boost judge performance in~\Cref{sec:inf-time} and~\Cref{app:structured-prompts}.

Generative judge models tend to lag specialized evaluators. MiniCheck naturally excels for faithfulness, while RAGAS offers more balanced, yet still competitive performance across refusal and faithfulness splits. However, most judges outperform the embedding-based RAGAS completeness score, showing an advantage of generative evaluation. 

Models tend to struggle with conciseness and unanswerable refusals. The difficulty with conciseness may be exacerbated by length bias~\citep{zeng2023evaluating}, as selecting shorter concise responses conflicts with the tendency of judge models to prefer longer ones. Likewise, struggling to select accurate refusals may be a special case of concreteness bias~\citep{park2024offsetbias}, as judges are biased towards substantive responses. Further analysis in~\Cref{app:judge-ft} reveals that poor accurate refusal performance may be an unintended result of judge finetuning. The same issue explains why models perform best on the answerable split. Evaluation trends show increasing difficulty with factuality, completeness, and conciseness, due to subtle distinctions in deeper levels of evaluation workflow and the bias toward longer responses.

\begin{takeaway}{blue}
Strong judge and reasoning models tend to perform the best on contextual evaluation.
\end{takeaway}

\subsection{How do judges handle criteria?}\label{sec:criteria-eval}
Our analysis thus far has been outcome driven: We have not verified that judges make correct judgments based on the specified criteria. Here, we conduct model-assisted verification on a subset of judge models that generate explanations: SFRJudge-8B,70B, Atla-Selene-8B, Self-taught-evaluator, and two Llama models. For all judgments with the correct outcome, we prompt 
GPT-4o to determine from the judge explanation if the judgment was decided by the correct criteria (Full prompt in~\Cref{app:criteria-eval-prompt}). From this, we compute a \textit{verified consistent accuracy}. In~\Cref{fig:critique_eval}, we plot the verified accuracies of each judge against its original accuracies, with the black dashed-line indicating the upper bound, where all correct responses use the right criteria. On average, verified accuracies tend to be 20 absolute percent lower than outcome-based accuracy, revealing that judges are using incorrect reasoning to reach correct outcomes. Refusals and faithfulness are generally determined for the correct reasons, whereas completeness and conciseness are not, further highlighting the challenges of evaluation in the contextual settings. A similar trend holds for the \textit{optimistic} variant of verified consistent accuracy, where we consider a sample to be correct if any of the two consistency runs (1) returns the correct outcome and (2) uses the correct reasoning. Verified optimistic accuracy tends to track better with unverified accuracy than the non-optimistic pair, with line of best fit coefficients of 0.99 and 0.86, respectively.

\begin{takeaway}{blue}
Judges use incorrect reasoning to arrive at the correct judgment across all splits
\end{takeaway}

\begin{figure*}[t]
    \centering
    \includegraphics[width=0.87\textwidth]{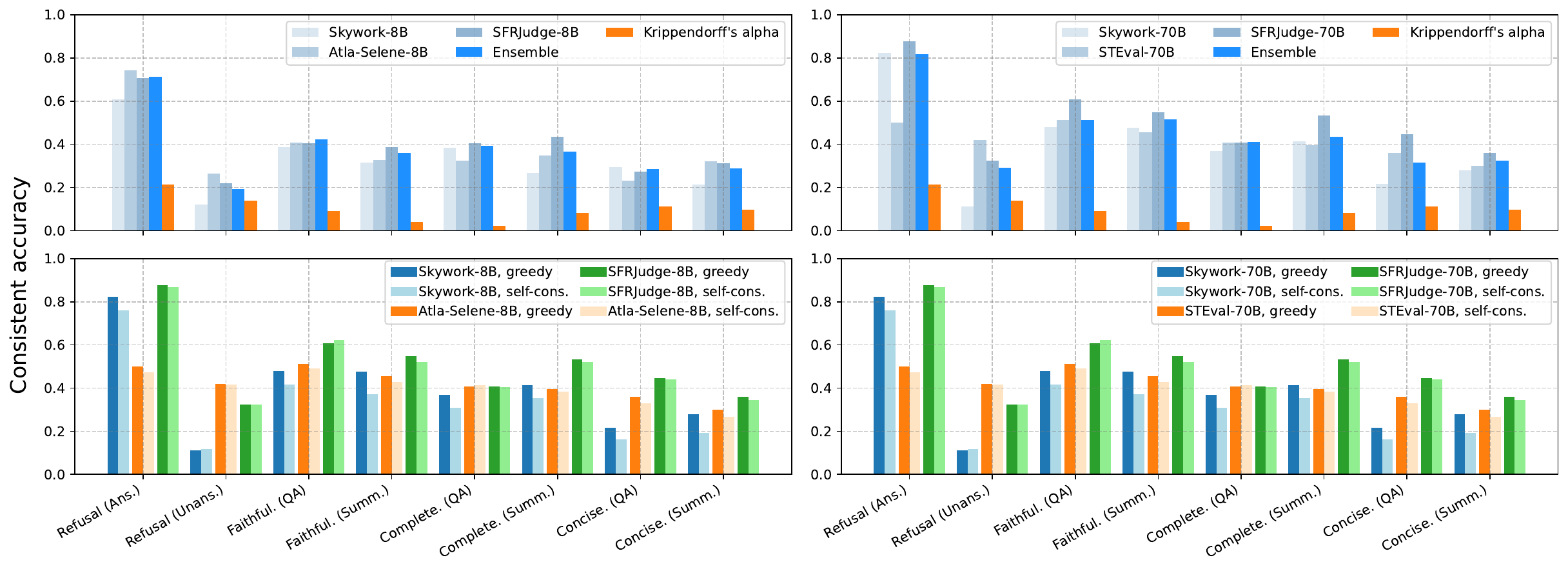}
    
    \caption{Across both small (Left) and larger (Right) judges, using inference time scaling has little effect. (Top) Ensembling judges into juries rarely outperforms the strongest judge in the jury due to weak judge agreement. (Bottom) Self-consistency rarely improves judge performance.}
    \label{fig:ensemble}
    
\end{figure*}

\begin{figure}
    \centering
    \includegraphics[width=0.87\columnwidth]{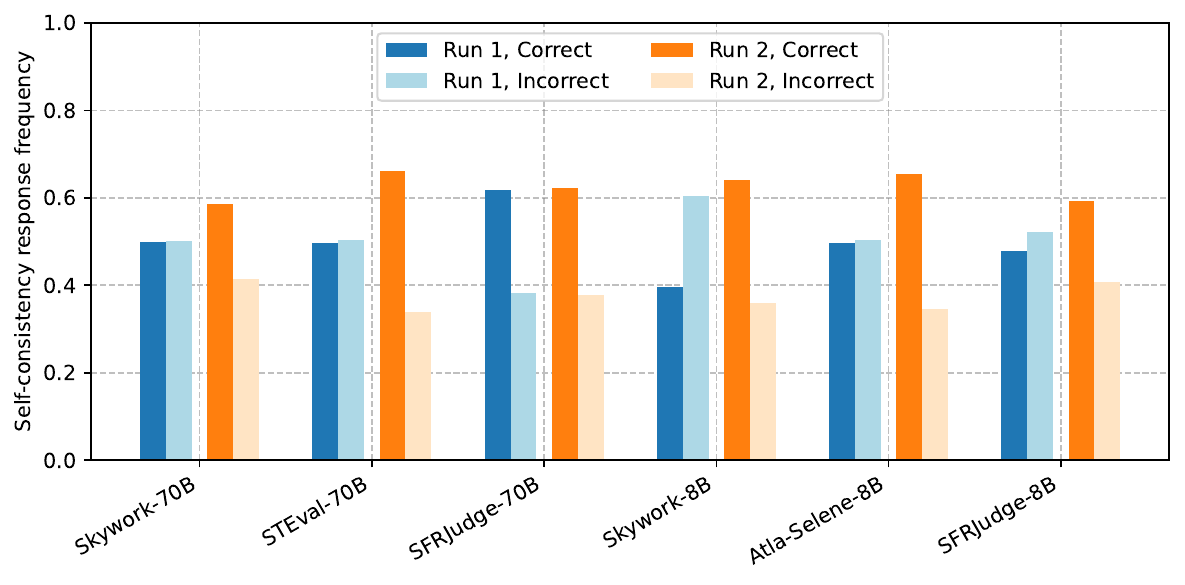}
    \caption{Distribution of judge responses by consistency run for $10$ sample self consistency. Each distribution is aggregated across splits.}
    \label{fig:self-cons-hist}
\end{figure}

While judges struggle to use the correct criteria when evaluating based on the contextual hierarchy, 
they are slightly more capable when given the correct criteria to use, as shown in~\Cref{fig:criteria-specific}. For each split, we prompt the judge with only the split criteria, omitting any mention of the contextual hierarchy. We compare judge performance against prompting with the full hierarchy. Conciseness and unanswerable refusals receive the greatest benefit, showing that length bias and concreteness bias can be mitigated to a degree with specific prompting.
However, performance gains are relatively muted across judges due to little change in judge consistency between the two settings. 
Judge inconsistency, even after abstracting away the hierarchical structure, suggests that contextual evaluation poses challenges beyond applying the correct criteria. Full results for all judge models are presented in~\Cref{app:criteria_specific}.

\begin{takeaway}{blue}
Judges are more capable when given the exact criteria to use, but are still inconsistent.
\end{takeaway}

\subsection{Can scaling inference-time compute help?}\label{sec:inf-time}
Inspired by recent efforts in inference-time scaling~\citep{jaech2024openai,snell2024scaling}, we investigate the impact of two test-time scaling techniques:  LLM-as-jury~\citep{verga2024replacing} and self-consistency~\citep{wang2022self}. We experiment with three smaller (8B) and three larger (70B) judges, and for both settings, aggregate judgments via majority vote\footnote{
    We treat inconsistent judgments as ties. For a sample, if the aggregated judgments do not have a clear winner, e.g., (A, Tie, B) or (Tie, Tie, Tie), then we consider it incorrect.
}. 
In~\Cref{fig:ensemble}, we present our results for both LLM-as-jury (top) using responses from different three judges and self-consistency (bottom) using 10 responses per judge (using a temperature of 0.7). The results are similar between smaller and larger models: LLM-as-jury rarely outperforms the strongest judge in the jury, while using self-consistency similarly has little impact on judge performance. 

\begin{figure}[t!]
    \centering
    \includegraphics[width=0.87\columnwidth]{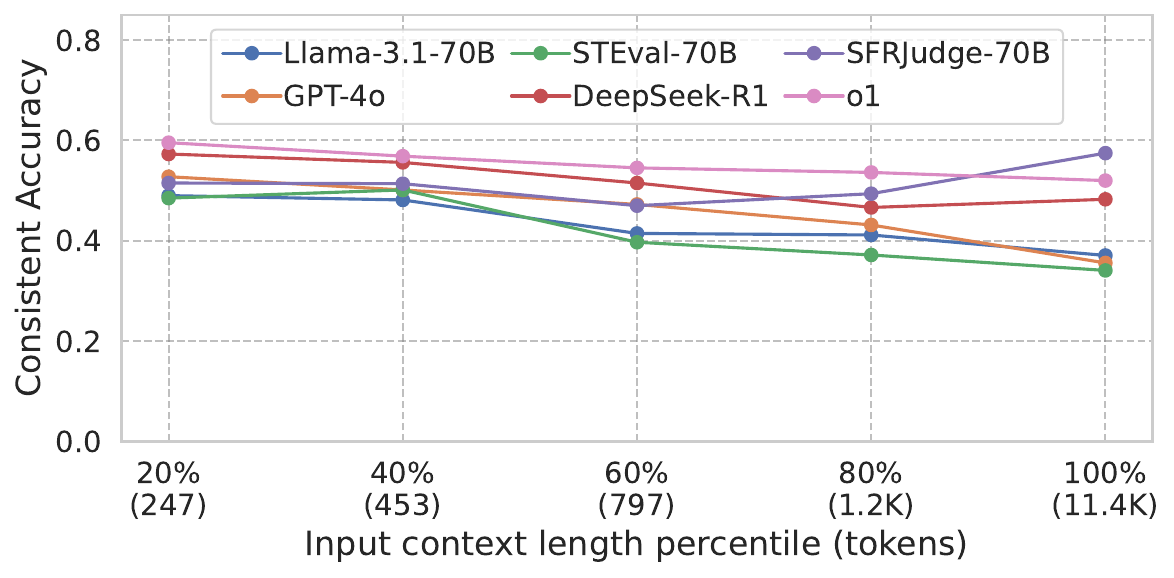}
    \caption{Judge performance decreases as context length increases. x-axis: Percentile of entire benchmark context lengths, with raw token count in parentheses. 
    }
     \label{fig:context_length}
\end{figure}

\begin{figure*}[ht!]
    \centering
    \includegraphics[width=0.87\textwidth]{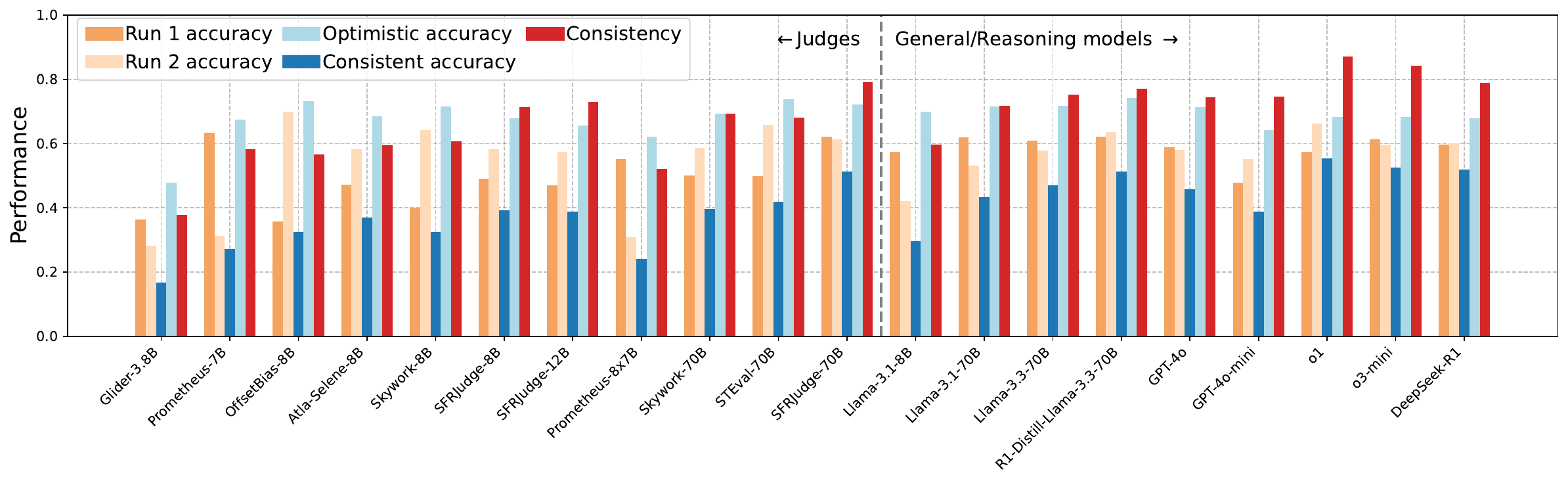}
    \vspace{-2mm}
    \caption{Four accuracy measures showing performance variations due to inconsistency, averaged across all splits for judge models (Left) and general purpose/reasoning models (Right).}
    \label{fig:bias-consistency}
\end{figure*}

\begin{figure*}[h!]
    \centering
    \includegraphics[width=0.87\columnwidth]{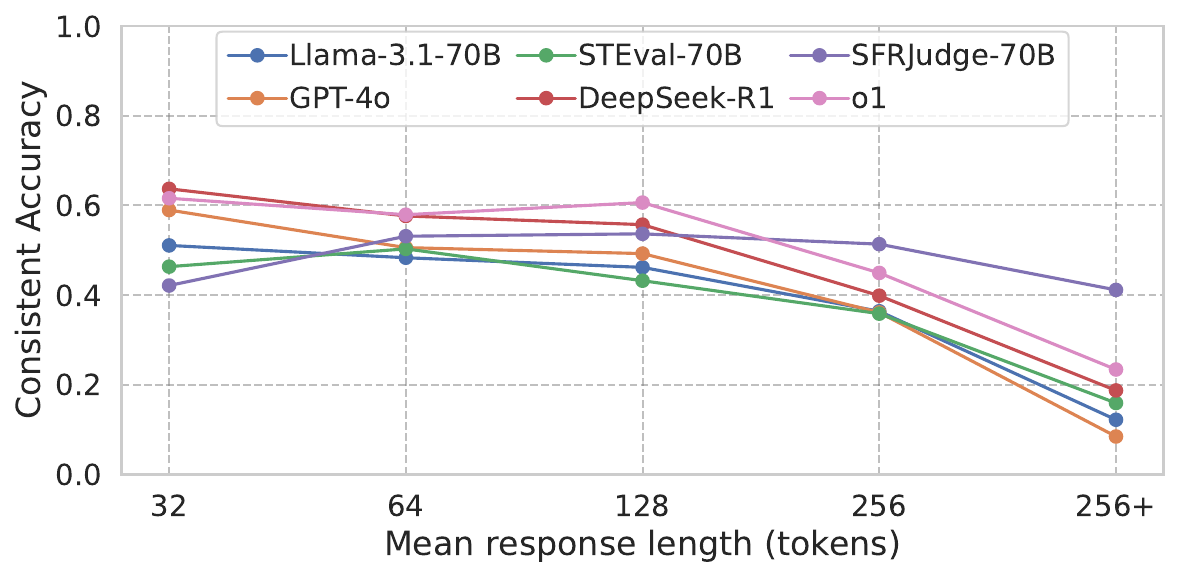}
    \includegraphics[width=0.87\columnwidth]{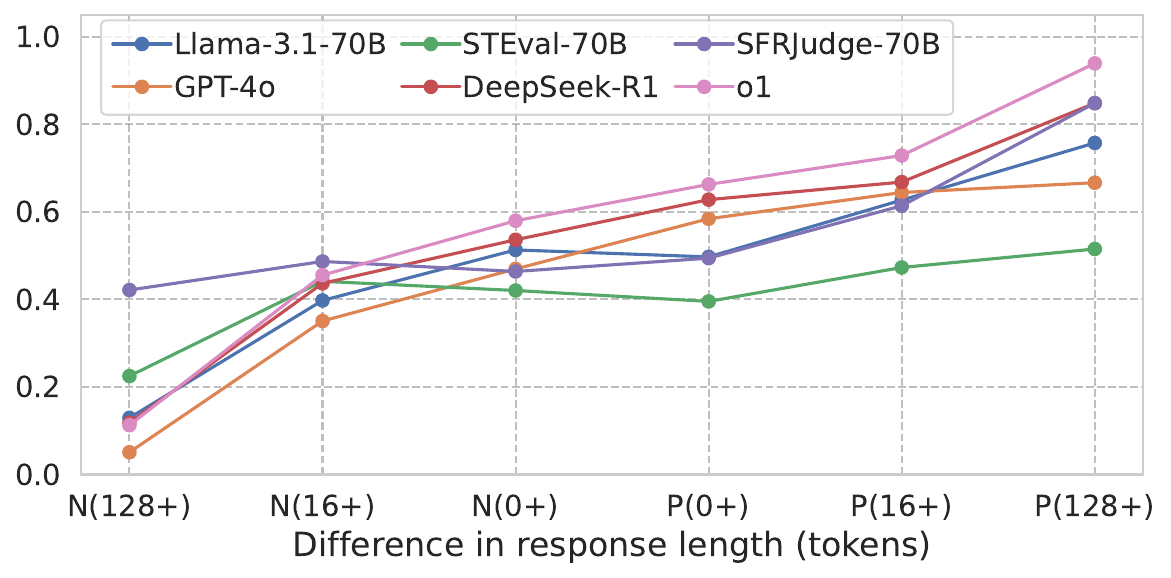} 
    \caption{
    Judge performance averaged over all splits as a function of two measures of response length. (Left): Judge performance decreases as response lengths increase. (Right): Judge performance increases as the difference in response length between the positive response and negative response grows, indicating a bias for longer responses. N(x+) and P(x+) indicate that the negative response and positive response is longer by x tokens or more, respectively.
    }
    \label{fig:model_response}
\end{figure*}

These trends may be surprising given the strong performance of reasoning models like o1 and DeepSeek-R1. The lack of jury success stems from the fact that judges do not exhibit \textit{structured} agreement. We use all judge outputs to compute Krippendorff's alpha coefficient~\citep{krippendorff2011computing}, which measures inter-annotator agreement on a range from -1 (complete disagreement) to 1 (complete agreement), with 0 indicating random chance. As shown in~\Cref{fig:ensemble}, judge agreement is extremely random: Even on the best-performing split, the alpha coefficient barely exceeds 0.2.

Lack of improvement from self-consistency likely results from the fact that contextual assessment is largely unseen in judge training. As a result, better judgments cannot be extracted via random sampling. To better visualize judge performance for self-consistency, we plot a histogram of the response distribution for each consistency run in~\Cref{fig:self-cons-hist}. Interestingly, the self-consistency judgment distribution may be the byproduct of \textit{positional bias}, where the judge outcome changes as a result of the order of responses in the prompt. Our findings in~\Cref{sec:pos-bias-intro} show that most judges exhibit slight positional preference for the second consistency run, with the lone exception of the positionally unbiased SFRJudge-70B. These trends are reflected in the self-consistency judgment distribution: The judgment distribution is more skewed towards correct responses in Run 2 compared to Run 1 for all judges except SFRJudge-70B. However, this does not translate to performance gains for SFRJudge-70B, as judgment pairs themselves may be inconsistent. 

\begin{takeaway}{blue}
    Common inference-time scaling approaches do not improve existing judge performance.
\end{takeaway}

%% file: figures/main_results.tex
\begin{table*}[ht]
    \centering
    \resizebox{0.98\textwidth}{!}{
    \begin{tabular}{c| lccccccccc}
%        \toprule
         & \multirow{2}{*}{Model} & Refusal & Refusal& Faithfulness & Faithfulness  & Completeness & Completeness & Conciseness & Conciseness  & \multirow{2}{*}{Average}  \\
        & & (Ans.) & (Unans.) & (QA) & (Summ.) & (QA) & (Summ.) & (QA) & (Summ.) & \\
        \midrule
        \multirow{7}{*}{\rotatebox{90}{Small Judge}}
        & Glider-3.8B & 12.0 & 8.8 & 45.6 & 9.2 & 20.8 & 28.7 & 5.1 & 4.1 & 16.8 \\
        & Promtheus-2-7b & 12.4 & 44.0 & 27.2 & 32.0 & 24.0 & 42.6 & 6.7 & 29.5 & 27.3 \\
        & Llama-3-OffsetBias-8B & 64.8 & 11.2 & 34.0 & 26.4 & 33.2 & 21.1 & 46.3 & 23.0 & 32.6 \\
        & Skywork-8B & 60.8 & 12.0 & 38.8 & 31.6 & 38.4 & 26.7 & 29.4 & 21.3 & 32.4 \\
        & Alta-Selene-8B & 74.4 & 26.4 & 40.8 & 32.8 & 32.4 & 34.7 & 23.1 & 32.0 & 37.1 \\
        & SFRJudge-8B & 70.8 & 22.0 & 40.4 & 38.8 & 40.4 & 43.4 & 27.5 & 31.1 & \textbf{39.3} \\
        & SFRJudge-12B & 68.4 & 28.4 & 45.2 & 43.6 & 28.0 & 51.0 & 16.1 & 29.5 & 38.8 \\
        \midrule
        \multirow{4}{*}{\rotatebox{90}{Large Judge}}        
        & Promtheus-2-8x7b & 22.0 & 29.6 & 22.4 & 29.6 & 20.4 & 39.8 & 10.2 & 18.4 & 24.1 \\
        & Skywork-70B & 82.4 & 11.2 & 48.0 & 47.6 & 36.8 & 41.4 & 21.6 & 27.9 & 39.6 \\
        & ST-Eval-70B & 50.0 & 42.0 & 51.2 & 45.6 & 40.8 & 39.4 & 36.1 & 29.9 & 41.9 \\
        & SFRJudge-70B & 87.6 & 32.4 & 60.8 & 54.8 & 40.8 & 53.4 & 44.7 & 36.1 & \textbf{51.4} \\
        \midrule       
        \multirow{9}{*}{\rotatebox{90}{Instruct + Reasoning}}
        & Llama-3.1-8B & 28.0 & 43.2 & 34.8 & 34.8 & 23.2 & 41.0 & 11.4 & 21.3 & 29.7 \\
        & Llama-3.1-70B & 59.6 & 48.0 & 58.0 & 48.4 & 38.0 & 51.8 & 15.7 & 27.5 & 43.4 \\
        & Llama-3.3-70B	& 71.6	& 42.4 & 68.0 &	48.4 & 42.0 &	51.8 &	20.8 &	30.7 & 47.0 \\
        & R1-Distill-Llama-3.3-70B	& 89.6 & 50.4 &	74.0 & 48.4 & 42.4 & 57.4 & 19.2 &	29.5 & 51.4 \\
        & GPT-4o-mini & 71.2 & 22.8 & 45.6 & 42.4 & 33.2 & 54.2 & 11.8 & 29.5 & 38.8 \\
        & GPT-4o & 64.0 & 52.0 & 68.0 & 50.8 & 39.6 & 56.2 & 12.9 & 22.5 & 45.8 \\
        & o3-mini & 95.2 & 34.4 & 76.4 & 58.0 & 40.4 & 59.8 & 20.8 & 35.7 & 52.6 \\
        & o1 & 96.0 & 48.4 & 84.4 & 59.2 & 48.4 & 63.7 & 15.3 & 27.0 & 55.3 \\
        & DeepSeek-R1 & 92.0 & 52.0 & 72.0 & 50.4 & 41.2 & 60.6 & 20.4 & 26.2 & 51.9 \\
        \midrule
        \multirow{2}{*}{\rotatebox{90}{Other}}
        & RAGAS & 62.4 & 60.0 & 78.8 & 54.4 & 22.4 & 23.1 & -- & -- & -- \\      
        % & Minicheck-7B (label) & 82.4 & 8.4 & 49.2 & 37.2 & -- & -- & -- & -- & --  \\
        & Minicheck-7B & 93.6 & 20.4 & 83.2 & 70.4 & -- & -- & -- & -- & --  \\
        \bottomrule
    \end{tabular}}
        \caption{\small Consistent accuracy for judge models, open-source instruct models, and API models on \name{}. %\shafiq{(consistent accuracy)}.%\austin{I'd remove average computation for RAGAS and minicheck, as they're not full avgs}
        }
        \label{tab:main-results}
\end{table*}

%% file: src/05Analysis.tex
\begin{figure*}[h!]
    \centering
    \includegraphics[width=0.95\columnwidth]{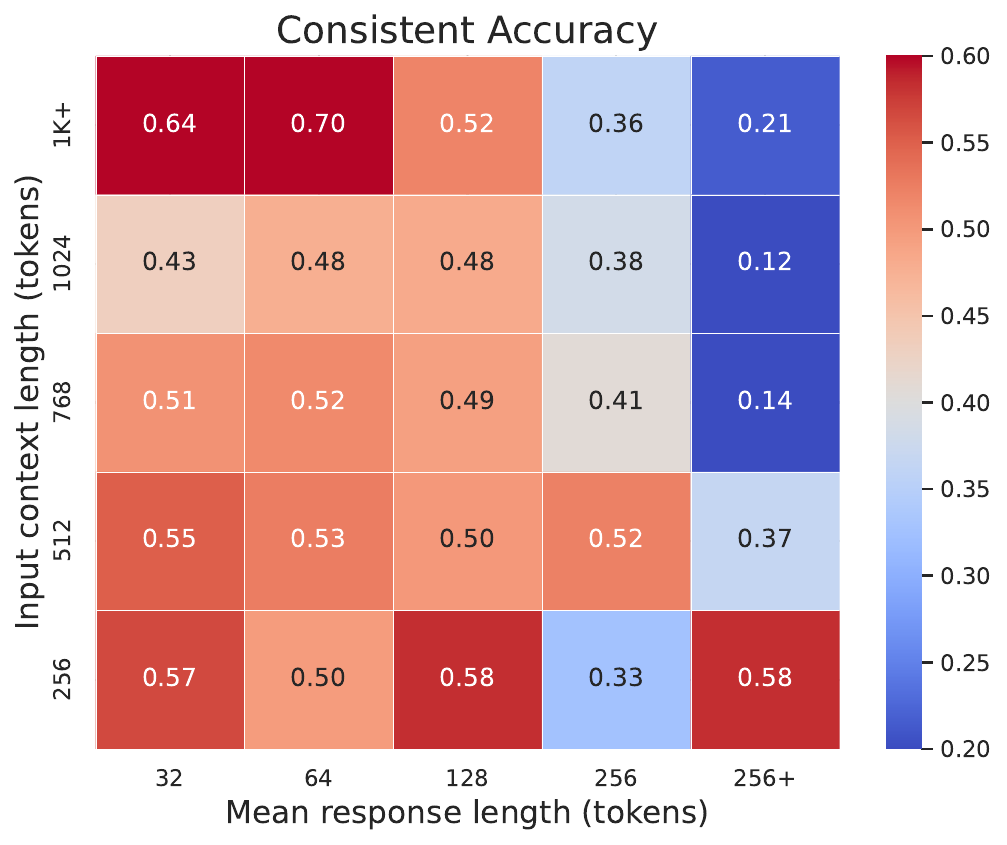}
    \includegraphics[width=0.95\columnwidth]{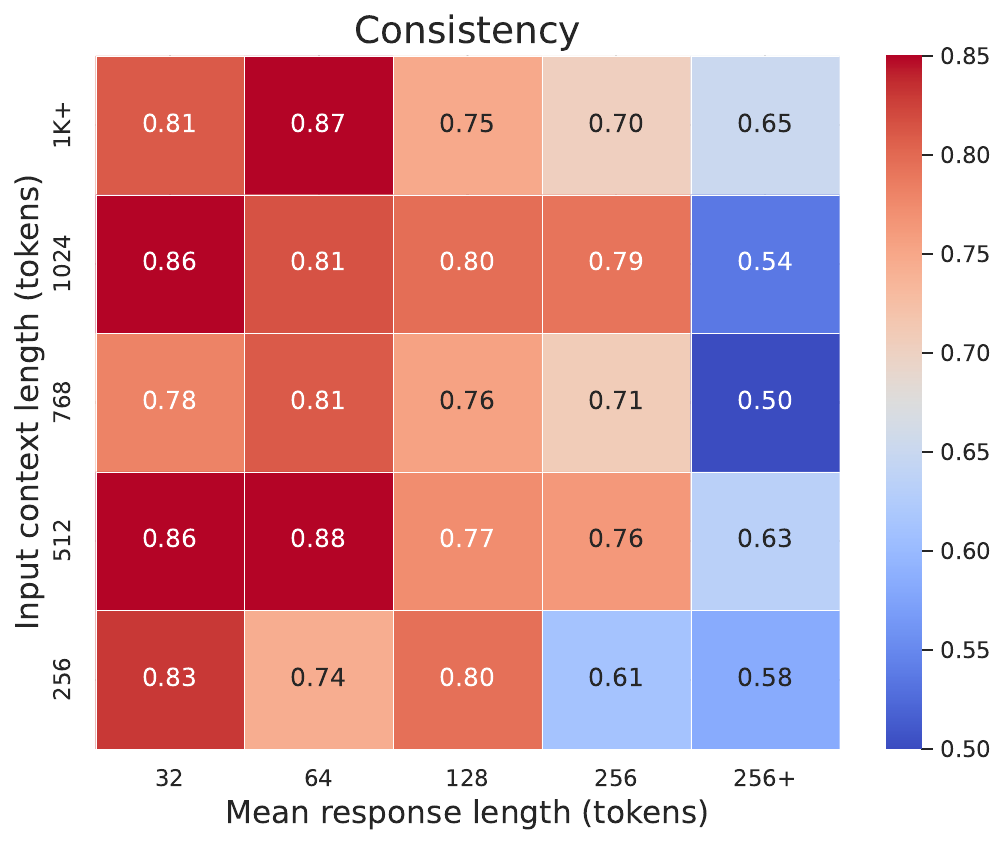} 
    \caption{Effects of length of both context and responses on judge accuracy (left) and consistency (right) averaged over the same subset of judges as in~\Cref{fig:context_length} and~\Cref{fig:model_response}.}
    \label{fig:heatmap}
\end{figure*}

\section{Bias Analysis and Discussion}
\subsection{How does input context affect judge performance?}\label{sec:bias:context_length}
The inclusion of context for judge models makes contextual evaluation more difficult: The longer the context, the more information the judge must comprehend before making a judgment. This section explores the interplay of context length and judge performance. In~\Cref{fig:context_length}, we visualize consistent accuracy as a function of input context tokens for a subset of strong performing judge models, binning context tokens by percentile. In general, judge performance decreases as the context length increases, with relatively weaker evaluators (e.g., GPT-4o) exhibiting larger relative drops than stronger evaluators (e.g., o1). Interestingly, SFRJudge-70B exhibits the most stable performance, with a slight increase in judge performance.
\begin{takeaway}{blue}
    Judge performance decreases as context length increases   
\end{takeaway}

\subsection{Do judges exhibit positional bias in contextual evaluation?}\label{sec:pos-bias-intro}
Past studies have noted that judges are not robust to the order of the response pairs~\citep{wang2023large,li2023generative}. This \textit{positional bias} may be further exacerbated by the inclusion of context. Inconsistency due to positional bias leads to performance variations for both judge models and general-purpose/reasoning models, as shown in~\Cref{fig:bias-consistency}. 
Here, we visualize the Run 1 and Run 2 accuracy along with consistent accuracy and optimistic accuracy, as defined in~\Cref{sec:eval-setup}.

The inter-run performance gap tends to be small for stronger models, such as larger finetuned judge models or reasoning models, reflecting more consistent judgments (higher consistency). In contrast, weaker models exhibit greater positional bias (lower consistency). The favored position varies by model. For example, Prometheus-7B and Llama models prefer the first response, while OffsetBias and OpenAI models favor the second. The optimistic accuracy shows that judges are not wrong in a consistent manner, but often change their judgments based on the order of responses. Notably, optimistic accuracy of finetuned judges is generally \textit{higher} than that of prompted judges (e.g., 73.1 for OffsetBias vs. 68.3 for o1), revealing that judge finetuning may raise the upper bound of evaluation. 

\begin{takeaway}{blue}
    Judge models are inconsistent rather than consistent and incorrect in contextual evaluation settings, revealing positional biases.
\end{takeaway}

\subsection{How does response length affect judge performance?}\label{sec:bias:response_length}

In additional to processing a potentially lengthy context, judge models also process the entirety of two responses. As response length increases, evaluation difficulty is expected to increase as well, as the judge is tasked with comparing and evaluating more content. This section examines how response length impacts judge performance. In~\Cref{fig:model_response} (Left), we plot consistent accuracy against the mean response length for a subset of strong judge models. Overall, performance declines as responses grow longer, with a similar trend across both relatively weak and strong evaluators. Similar to trends with increasing context length (\Cref{sec:bias:context_length}), SFRJudge-70B remains the most stable, showing minimal performance variation.

\begin{takeaway}{blue}
    Judge performance is inversely correlated with response length: Judges are more accurate for shorter responses.
\end{takeaway}

Beyond the absolute length of responses, the relative difference in the length between Response A and Response B can also impact judge performance. It has been widely noted in prior work that judges exhibit response length bias~\citep{zeng2023evaluating,park2024offsetbias}, i.e., judges prefer longer responses, even if said responses are not meaningfully better.~\Cref{fig:model_response} (Right) plots judge performance as a function of the \textit{difference} in length between positive and negative responses. Judges tend to struggle to identify the positive response if the negative response is longer by significant (e.g., 128+ tokens) and excell at identifying the correct response when the positive response is longer. This overall trend indicates that judges are biased towards longer responses, which may partially explain the relatively low performance of judge models on the conciseness splits As reported in~\Cref{tab:dataset_splits}, positive positive responses are \textit{shorter} than negative responses on average, meaning that a high-performing judge must fight against its inherent bias towards longer responses to select the better response.

\begin{takeaway}{blue}
    Judges exhibit response length bias, which may impact performance when using length-correlated criteria, like conciseness.
\end{takeaway}

\subsection{How do context and response length compound evaluation difficulty?}\label{sec:pos-bias}
Thus far, we've considered how the length of context and responses have affected judge performance in isolation. In~\Cref{fig:heatmap}, we visualize at their combined affect on consistent accuracy (left) and judge consistency (right), a measure of positional bias for the same subset of judges as the previous two sections. In general, response length and context length tend to have a compounding effect: As responses and context both increase in length, judge accuracy and consistency tend to decrease, with longer responses impacting performance more than longer input context in general. The opposite trend generally holds as well, with stronger judge performances coming with shorter contexts and/or responses. 

%% file: src/06Conclusion.tex
\section{Conclusion}
We introduce \name{}, a benchmark designed to evaluate LLM-judges in contextual settings. Building on a principled contextual evaluation hierarchy, we construct eight benchmark splits that assess refusals, faithfulness, completeness, and conciseness. This benchmark presents a significant challenge for state-of-the-art judge and reasoning models, with SFRJudge-70B and o1 achieving consistent accuracies of 51.4\% and 55.3\%, respectively. Additionally, we conduct a thorough analysis of reasoning correctness and examine the impact of common methods for scaling test-time compute, result of which further validate the unique challenges of contextual evaluation. 

%% file: src/10Appendix.tex
\section{\name{}: Additional details} \label{app:add_details}

\input{figures/data_examples}
We provide detailed information on all eight splits of \name{}, covering the datasets utilized, the applied methodologies, and the prompts used for data generation. For a concise summary, refer to Table \ref{tab:data_approaches}, and for split-specific examples, see Table \ref{tab:app-examples}.

We sampled data using a model-based response perturbation approach, as introduced in \Cref{sec:benchmark}, which includes two methods: desired output prompting (\textbf{M1}) and existing output modification (\textbf{M2}).\\

\noindent\textbf{Prompt 1 (M1)}: Generate context-based refusal \\
\textbf{Prompt 2 (M1)}: Generate hallucinated response \\
\textbf{Prompt 3 (M1)}: Generate unfaithful response \\
\textbf{Prompt 4 (M2)}: Generate less-complete response \\
\textbf{Prompt 5 (M2)}: Generate less-concise response

\begin{tcolorbox}[breakable,enhanced, left=-0.5cm, right=-0.5cm, top=2pt, bottom=2pt, enlarge top by=0.1cm, enlarge bottom by=0.1cm]
\begin{quote}
\begin{lstlisting}
### Data Prompt 1 (M1) : Generate context-based refusal response

Given the following question and context, generate a context-specific refusal response that:
1. Explains why you cannot answer based on the given context
2. Points out specific missing information or topics that would be needed to answer properly
3. May suggest what kind of information would be needed to provide a proper answer
4. Avoids generic responses like "I don't know" or "not enough information"

Examples of good refusals:
- "Although the context discusses cooking temperatures, it lacks crucial information about high-altitude adjustments and humidity factors that would be necessary to properly answer this baking question."
- "While the context mentions flight regulations, it doesn't include specific airline policies or statistical data on baggage handling that would be needed to fully address this question."
- "The context touches on basic gardening principles but lacks essential details about soil pH levels and regional climate variations needed to answer this plant care question."

Important: Even if the context contains partial information, explain what specific information you claim is missing.

Question: 
{question}

Contexts:
{context}

Please provide:
1. A convincing refusal response that explains what specific information you claim is missing
2. A brief justification explaining why this refusal appears appropriate
     
Format the response as a JSON object with keys 'refusal_response' and 'justification'.
\end{lstlisting}
\end{quote}
\label{prompt:p1}
\end{tcolorbox}

\begin{tcolorbox}[breakable,enhanced, left=-0.5cm, right=-0.5cm, top=2pt, bottom=2pt, enlarge top by=0.1cm, enlarge bottom by=0.1cm]
\begin{quote}
\begin{lstlisting}
### Data Prompt 2 (M1) : Generate hallucinated response

Given the following question and context, provide a well-thought, and specific answer:

Question: 
{question}

Contexts:
{context}
    
Please provide:
1. A step-by-step reasoning process explaining how you arrive at your answer
2. A final, direct answer based on this reasoning
    
You must provide a specific answer. You cannot respond with "I don't know" or "not enough information".
    
Format the response as a JSON object with two keys:
- 'thoughts': Your step-by-step reasoning process
- 'answer': Your final answer.
\end{lstlisting}
\end{quote}
\label{prompt:p2}
\end{tcolorbox}

\begin{tcolorbox}[breakable,enhanced, left=-0.5cm, right=-0.5cm, top=2pt, bottom=2pt, enlarge top by=0.1cm, enlarge bottom by=0.1cm]
\begin{quote}
\begin{lstlisting}
### Data Prompt 3 (M1) : Generate unfaithful response

Given the following question and its faithful answer, generate an unfaithful answer (unfaithful with respect to the given context) that:
1. Can be a correct answer to the question.
2. May include plausible-sounding but irrelevant information with respect to the given contexts.
Question: 
{question}

Contexts:
{context}

Faithful Answer:
{answer}

Please provide:
1. An unfaithful answer
2. A brief justification explaining why the answer is unfaithful (irrlevant) to the context.
Format the response as a JSON object with keys 'unfaithful_answer' and 'justification'.
\end{lstlisting}
\end{quote}
\label{prompt:p3}
\end{tcolorbox}

\begin{tcolorbox}[breakable,enhanced, left=-0.5cm, right=-0.5cm, top=2pt, bottom=2pt, enlarge top by=0.1cm, enlarge bottom by=0.1cm]
\begin{quote}
\begin{lstlisting}
### Data Prompt 4 (M2) : Generate less-complete response
Task: Modify the given response by removing key details from one or more cited passages while maintaining a similar length by expanding on less relevant details.

Instructions:
1. Omit one or more cited passages to make the response less complete, removing essential details.
2. Compensate for the missing information by elaborating on other cited passages with unnecessary or redundant details.
3. Ensure the modified response remains factually accurate and aligns with the provided context.
4. Maintain a similar length to the original response, ensuring the new version differs by more than 10-15 words.
5. Avoid copying the structure of the given response; create a unique structure instead. \n6. Do not include citations (e.g., [*]) in the modified response. 

Question:
{question}

Context:
{context}

Response with citations:
{answer}
\end{lstlisting}
\end{quote}
\label{prompt:p4}
\end{tcolorbox}

\begin{tcolorbox}[breakable,enhanced, left=-0.5cm, right=-0.5cm, top=2pt, bottom=2pt, enlarge top by=0.1cm, enlarge bottom by=0.1cm]
\begin{quote}
\begin{lstlisting}
### Data Prompt 5 (M2) : Generate less-concise response

Task: Given the following question, context, and answer with citations, your task is to generate a less consise and more detailed response by expanding some of the citations through direct quotations from the cited passages. The response should include all relevant details form the original answer but should be rephrased to avoid copying directly. By incorporating specific lines from the cited articles, the response will become more authoritative. Not all citations need to expanded-choose which ones to elaborate on for the greatest impact. Ensure that the final response does not exceed the original length by more than 50 words and maintains a unique structure while conveying the same information. Do not include citations in the generated response.

Question:
{question}

Context:
{context}

Response with citations:
{answer}
\end{lstlisting}
\end{quote}
\label{prompt:p5}
\end{tcolorbox}

\section{Judge model details}\label{app:judge-models}
Here, we provide additional details about evaluated judge models, prompts used for judge models, and prompts used for model-assisted criteria evaluation.

\subsection{Overview of judge model baselines}\label{app:judge-model-details}
We evaluate the 11 judge models from the following judge families.
\begin{itemize}[leftmargin=*,noitemsep,topsep=5pt]
    \item \textbf{GLIDER} \citep{deshpande2024glider}: GLIDER is finetuned from Phi-3.5-mini-instruct~\citep{abdin2024phi} to be a lightweight evaluator. GLIDER is trained with anchored preference optimization~\citep{d2024anchored} to perform pairwise, single-rating, and binary classification evaluation, while producing explanations.
    \item \textbf{Prometheus-2} \citep{kim2024prometheus}: The Prometheus-2 family of models are finetuned from Mistral 7B and 8x7B instruct models~\citep{jiang2023mistral,jiang2024mixtral} to conduct pairwise and single-rating evaluation. They utilize purely synthetic data distilled from GPT-4 to train their models to produce explanations and judgments.
    \item \textbf{OffsetBias} \citep{park2024offsetbias}: OffsetBias is finetuned from Llama-3-Instruct~\citep{dubey2024llama} to perform pairwise comparison evaluation. It is trained with supervised finetuning (SFT) explicitly with an emphasis on bias mitigation via adversarially generated data. OffsetBias does not produce explanations.
    \item \textbf{Atla-Selene} \citep{alexandru2025atla}: Atla-Selene is a general purpose evaluator trained from Llama-3.1-8B instruct. It is trained to perform pairwise, single-rating, and binary classification evaluation via iterative reasoning preference optimization~\citep{pang2024iterative}.
    \item \textbf{Skywork-Critic} \citep{skyworkcritic2024}: Skywork-Critic judges are finetuned from Llama-3.1-8B and 70B instruct to perform pairwise evaluation. The emphasis of Skywork is in data curation, using a relatively small set judgments to train an evaluator with SFT. Skywork-Critic models do not generate explanations.
    \item \textbf{SFRJudge} \citep{wang2024direct}: SFRJudge are a family of judges finetuned from Mistral-NeMo-12B~\citep{mistralnvidia2024nemo} and Llama-3.1-8B and 70B instruct models to perform pairwise, single-rating, and binary classification evaluation. These models are trained with direct preference optimization~\citep{rafailov2024direct} with an emphasis on training tasks. SFRJudge models are able to generate explanations.
    \item \textbf{Self-taught-evaluator} \citep{wang2024self}: Self-taught-evaluator is trained form Llama-3.1-70B instruct using an iterative DPO training approach. This model is trained to produce explanations and conduct pairwise evaluation.
\end{itemize}

\subsection{Sample judge model prompt template}\label{app:judge-prompts}
For all judges, we preserve the model-developer provided template. This informs the judge of expected data format and corresponding output format. We additionally use provided judgment parsing code when available. We utilize the same evaluation description across all judges. We present full prompt examples below for our standard prompt, which describes the entire workflow, our structured prompt, which emphasizes faithfulness via structured chain-of-thought (as discussed in~\Cref{app:structured-prompts}), and our criteria-specific prompts used in~\Cref{sec:criteria-eval}.

\begin{tcolorbox}[breakable,enhanced, left=-0.5cm, right=-0.5cm, top=2pt, bottom=2pt, enlarge top by=0.1cm, enlarge bottom by=0.1cm]
\begin{quote}
\begin{lstlisting}
### Standard prompt

You are a contextual judge. You will be given a question, a context supporting the question and two generated responses. Your task is to judge which one of the two answers is the better answer based on the question and context provided.
Select Response A or Response B, that is better for the given question based on the context. The two responses are generated by two different AI chatbots respectively.
Do NOT say both / neither are good.

Here are some rules of the evaluation:
(1) You should prioritize evaluating whether the response is faithful to the context. A response is faithful to the context if all of the factual information in the response is attributable to the context. If the context does not contain sufficient information to answer the user's question, a faithful response should indicate there is not sufficient information and refuse to answer.
(2) You should pick the response that is more faithful to the context.
(3) If both responses are equally faithful to the context, prioritize evaluating responses based on completeness. A response is complete if it addresses all aspects of the question.
If two responses are equally complete, evaluate based on conciseness. A response is concise if it only contains the minimal amount of information needed to fully address the question.
(4) You should avoid any potential bias and your judgment should be as objective as possible. Here are some potential sources of bias:
- The order in which the responses were presented should NOT affect your judgment, as Response A and Response B are **equally likely** to be the better.
- The length of the responses should NOT affect your judgement, as a longer response does not necessarily correspond to a better response. When making your decision, evaluate if the response length is appropriate for the given instruction.

Your reply should strictly follow this format:
**Reasoning:** <feedback evaluating the responses>

**Result:** <A or B>

Here is the data.

Question:
```
{question}
```

Response A:
```
{response_a}
```

Response B:
```
{response_b}
```

Context:
```
{context}
```
\end{lstlisting}
\end{quote}
\end{tcolorbox}

\begin{tcolorbox}[breakable,enhanced, left=-0.5cm, right=-0.5cm, top=2pt, bottom=2pt, enlarge top by=0.1cm, enlarge bottom by=0.1cm]
\begin{quote}
\begin{lstlisting}
### Structured prompt

You are a contextual judge. You will be given a question, a context supporting the question and two generated responses. Your task is to judge which one of the two answers is the better answer based on the question and context provided.
Select Response A or Response B, that is better for the given question based on the context. The two responses are generated by two different AI chatbots respectively.
Do NOT say both / neither are good.

Here are some rules of the evaluation:
(1) A response is faithful to the context if all of the factual information in the response is attributable to the context. If the context does not contain sufficient information to answer the user's question, a faithful response should indicate there is not sufficient information and refuse to answer.
(2) First, determine if Response A is faithful to the context. Provide reasoning for your decision, then write your response as **Response A:** <yes/no>
(3) Second, determine if Response B is faithful to the context. Provide reasoning for your decision, then write your response as **Response B:** <yes/no>
(4) If one response is faithful while the other response is not, select the faithful response. If both responses are equally faithful to the context, prioritize evaluating responses based on {criteria}.
(5) You should avoid any potential bias and your judgment should be as objective as possible. Here are some potential sources of bias:
- The order in which the responses were presented should NOT affect your judgment, as Response A and Response B are **equally likely** to be the better.
- The length of the responses should NOT affect your judgement, as a longer response does not necessarily correspond to a better response. When making your decision, evaluate if the response length is appropriate for the given instruction.

Your reply should strictly follow this format:
**Response A reasoning:** <reasoning for response A faithfulness>

**Response A:** <yes/no if response A is faithful to the context>

**Response B reasoning:** <reasoning for response B faithfulness>

**Response B:** <yes/no if response B is faithful to the context>

**Reasoning:** <feedback evaluating the responses>

**Result:** <A or B>

Here is the data.

Question:
```
{question}
```

Response A:
```
{response_a}
```

Response B:
```
{response_b}
```

Context:
```
{context}
```
\end{lstlisting}
\end{quote}
\end{tcolorbox}

Below is our criteria-specific prompts, where for criteria, we substitute in one of the following:
\begin{itemize}[leftmargin=*,noitemsep,topsep=5pt]
    \item \textbf{Refusal:} ``refusal validity. A response correctly refuses to answer if the context does not contain sufficient information to answer the user's question.''
    \item \textbf{Faithfulness:} ``faithfulness.  A response is faithful to the context if all of the factual information in the response is attributable to the context.''
    \item \textbf{Completeness:} ``completeness. A response is complete if it addresses all aspects of the question.''
    \item \textbf{Conciseness:} ``conciseness. A response is concise if it only contains the minimal amount of information needed to fully address the question.''
\end{itemize}

\begin{tcolorbox}[breakable,enhanced, left=-0.5cm, right=-0.5cm, top=2pt, bottom=2pt, enlarge top by=0.1cm, enlarge bottom by=0.1cm]
\begin{quote}
\begin{lstlisting}
### Criteria specific

You are a helpful assistant in evaluating the quality of the responses for a given instruction and context. Your goal is to select the best response for the given instruction and context.
Select Response A or Response B, that is better for the given instruction. The two responses are generated by two different AI chatbots respectively.
Do NOT say both / neither are good.

Here are some rules of the evaluation:
(1) You should prioritize evaluating on {criteria}
(2) Responses should NOT contain more/less than what the instruction asks for, as such responses do NOT precisely execute the instruction.
(3) You should avoid any potential bias and your judgment should be as objective as possible. Here are some potential sources of bias:
- The order in which the responses were presented should NOT affect your judgment, as Response A and Response B are **equally likely** to be the better.
- The length of the responses should NOT affect your judgement, as a longer response does not necessarily correspond to a better response. When making your decision, evaluate if the response length is appropriate for the given instruction.

Your reply should strictly follow this format:
**Reasoning:** <feedback evaluating the responses>

**Result:** <A or B>

Here is the data.

Question:
```
{question}
```

Response A:
```
{response_a}
```

Response B:
```
{response_b}
```

Context:
```
{context}
```
\end{lstlisting}
\end{quote}
\end{tcolorbox}

\begin{tcolorbox}[breakable,enhanced, left=-0.5cm, right=-0.5cm, top=2pt, bottom=2pt, enlarge top by=0.1cm, enlarge bottom by=0.1cm]
\begin{quote}
\begin{lstlisting}
### GPT-4o criteria evaluation prompt

You are given an <evaluation explanation>, a <evaluation outcome>, and a set of <criteria>. 
Another large language model conducted a pairwise evaluation between two responses, Response A and Response B. 
Based on the content of the <evaluation explanation>, your task is to decide if the <evaluation outcome> was decided based on <criteria>.
The <evaluation explanation> is allowed to mention criteria other than <criteria>. But it must use <criteria> as the primary criteria in its decision.

<evaluation explanation>: {critique}
<evaluation outcome>: {judgment}
<criteria>: {criteria}

Please give a short explanation, then respond with Yes or No. Use the format
<explanation>: your explanation
<decision>: Yes or No
\end{lstlisting}
\end{quote}
\end{tcolorbox}

\begin{figure*}[t]
    \centering
    \includegraphics[width=0.87\textwidth]{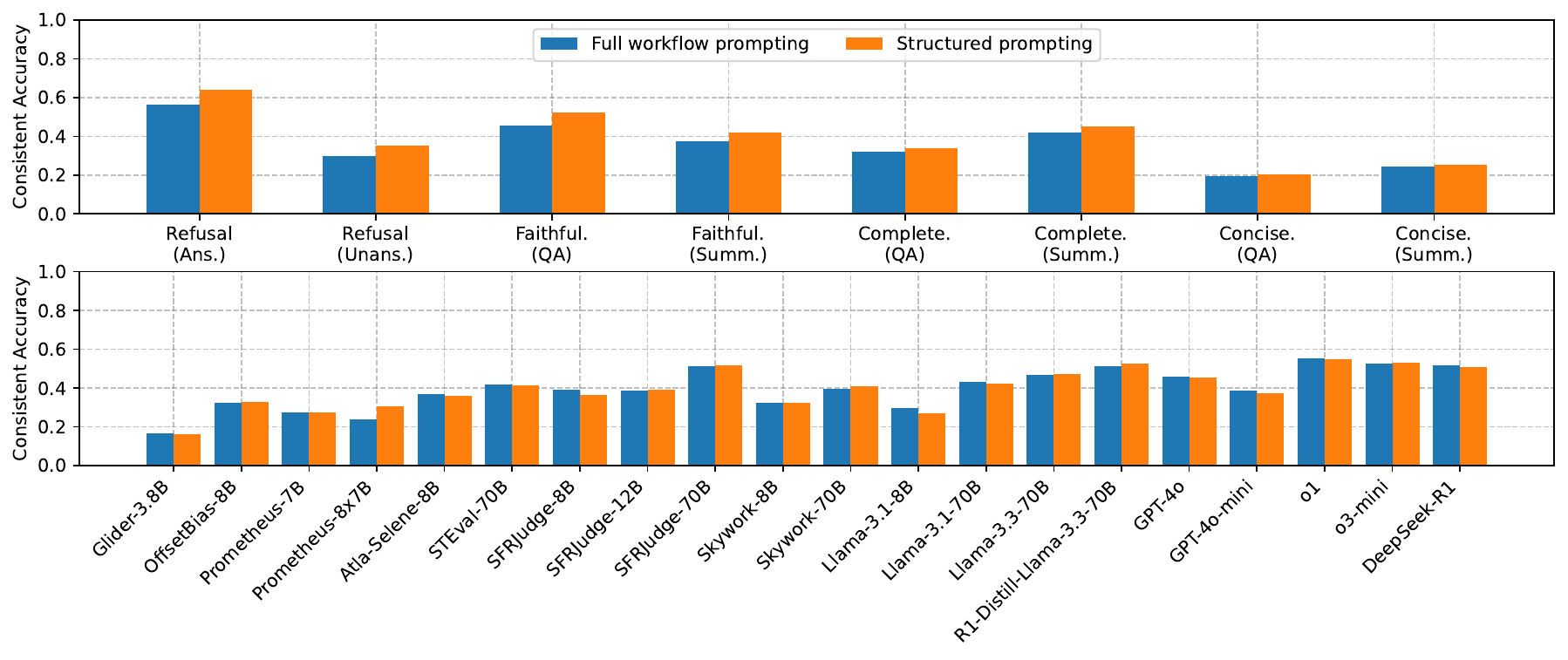}
    \caption{Using a structured chain-of-thought prompt by instructing judges to explicitly list out faithfulness evaluation before evaluating on other criteria does not lead to meaningful performance changes.}
    \label{fig:structured-prompts}
\end{figure*}
\begin{figure}[t]
    \centering
    \includegraphics[width=\columnwidth]{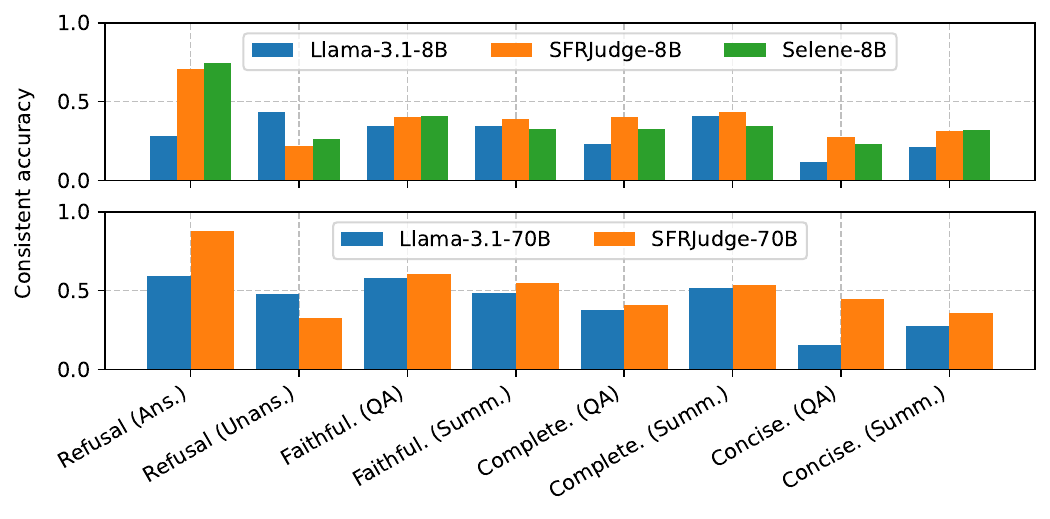}
    \caption{Non-contextual judge finetuning helps most splits relative to base model performance, but notably hurts unanswerable refusals.}
    \label{fig:judge-ft}
\end{figure}

\subsection{Criteria verification GPT-4o prompt}\label{app:criteria-eval-prompt}
Here, we present the prompt used for criteria verification in~\Cref{sec:criteria-eval}. For each split, we prompt GPT-4o to evaluate the response based on the judge explanation, judge output, and split criteria. For the criteria field, we use the following descriptions:

\begin{itemize}[leftmargin=*,noitemsep,topsep=5pt]
    \item \textbf{Refusal:} ``Refusal validity / faithfulness: The chosen response either correctly refuses to answer or correctly does not refuse and answers. This makes the chosen response appropriately faithful.''
    \item \textbf{Faithfulness:} ``Faithfulness: The chosen response is more faithful, factual, or truthful.''
    \item \textbf{Completeness:} ``Completeness: The chosen response is more complete, thorough, or comprehensive.''
    \item \textbf{Conciseness:} ``Conciseness: The chosen response is more concise or less wordy or verbose.''
\end{itemize}

\input{figures/criteria_specific_results}

\input{figures/app-examples}

\section{Additional experimental results}
\subsection{Can we improve performance with structured prompting?}\label{app:structured-prompts}

Our results in~\Cref{sec:main-results} reveal that judges struggle with verifying factuality, a key step early on in the evaluation workflow. Here, we experiment with a prompt (presented in~\Cref{app:judge-prompts}) that emphasizes factuality via a more structured output format. For judges that produce explanations, we ask the judge to determine each response's faithfulness independently, requiring it to output ``Response \{A,B\} faithfulness reasoning: <reasoning>'' and ``Response \{A,B\} faithfulness: <yes/no>'' before its evaluation along other workflow criteria. This can be viewed as directing the judge to produce a more structured chain-of-thought~\citep{li2025structured} before evaluation or using user-specified test cases~\citep{saad2024lmunit,saha2025learning}. For judges that do not produce explanations, we omit the reasoning requirement. We visualize the performance per-judge and per-split in~\Cref{fig:structured-prompts}, which reveals that structured prompting has minimal effects. Despite the prompt focus on factuality, performance in factuality splits only increases marginally. Performance shifts in either direction are minimal across most judges, with the out-of-training-distribution nature of this prompt likely offsetting any potential gains. As such, clever prompting at inference time cannot dramatically improve judge performance. %\shafiq{we can send this to App since it has minimal effect. But we should somewhere mention and refer the reader.}

\subsection{What does non-contextual judge finetuning help?}\label{app:judge-ft}
Judge models are typically finetuned starting from general-purpose instruct models. Here, we analyze the effects of such finetuning by comparing the SFRJudge-8B and Atla-Selene-8B to their original base model, Llama-3.1-8B, and SFRJudge-70B to Llama-3.1-70B. All models use the same prompt template for evaluation. As we visualize in~\Cref{fig:judge-ft}, judge finetuning for non-contextual evaluation still helps evaluation performance for most splits, but notably \textit{hurts} performance for identifying accurate refusals. This performance degradation may reveal one hidden assumption in judge model training: That the responses evaluated \textit{always} attempt to satisfy the user requests. That is, judge training data likely does not include examples of accurate refusals, leading to skewed performance for refusals, with large boosts in identifying inaccurate refusals, but sizable drops in identifying accurate refusals. This same trend holds for larger judge models too, albeit with slightly smaller changes in performance. This indicates that larger base models come with a higher level of ``fundamental judgment ability'' than smaller models, resulting in less gains from judge-specific training. However, this does not mean there are no tangible benefits, as highlighted in the increase in Conciseness (QA) performance.

\subsection{Complete experimental results for criteria-specific prompting}\label{app:criteria_specific}
In~\Cref{tab:criteria-specific-results} we present full results for criteria-specific prompting results presented in~\Cref{sec:criteria-eval}. Judge performance tends to increase slightly with fully-specified criteria, indicating that the full contextual hierarchy makes evaluation more difficult. However, results are still relatively low, as consistency does not increase significantly. This, compounded with the analysis presented in~\Cref{sec:pos-bias}, indicate that the additional context poses has unique challenges. Of judge models evaluated, Llama-3.1-8B notably exhibits extremely skewed performance in evaluating refusals: It is unable to correctly identify a incorrect refusal, preferring the refusal across both splits regardless of if the question is answerable or not.

%% file: figures/data_examples.tex
\begin{table*}[th]
    \centering
    \resizebox{0.95\textwidth}{!}{
    \begin{tabular}{p{0.8in}p{1.9in}p{3in}p{3in}}
        \toprule
        \textbf{Split} & \textbf{Dataset} & \textbf{Positive Response (approach)} & \textbf{Negative Response (approach)} \\
        \midrule
        \textbf{Refusal (Ans.)} & LFRQA\citep{han2024rag} & Provided response & Context-based refusal using data prompt-1 (\textbf{M1}) \\
        \midrule
        \textbf{Refusal (Unans.)} & FaithEval\citep{ming2024faitheval} & Context-based refusal using data prompt-1 (\textbf{M1}) &  Generate substantive response with data prompt-2 (\textbf{M1})\\
        \midrule
        \textbf{Faithfulness} & LFRQA\cite{han2024rag} & Provided response & Generate unfaithful response using data  \\
        \textbf{(QA)} & LFQA\citep{xu-etal-2023-critical} &  &  prompt-3 (\textbf{M1})\\
        & MRQA\cite{fisch-etal-2019-mrqa} &  &  \\
        
        & QA-Feedback\citep{wu2023finegrainedhumanfeedbackgives} & Faithful responses (\textbf{H}) & Unfaithful responses (\textbf{H}) \\
        & RAGTruth\citep{niu2023ragtruth} &  & \\

        \midrule
        \textbf{Faithfulness} & FineSumFact\citep{oh-etal-2025-learning} & Fully faithful responses or response with higher  & Unfaithful response with lower faithfulness \\
        \textbf{(Summ.)}& InstruSum\citep{liu-etal-2024-benchmarking} & faithfulness (0.75 or more) (\textbf{H}) & score (\textbf{H})\\
        & LongformFact\citep{wan2024positionalbiasfaithfulnesslongform} & & \\
        & UniSumEval\citep{lee-etal-2024-unisumeval} & & \\
        & FineSurE\citep{song-etal-2024-finesure} & & \\
        & RAGTruth \citep{niu2023ragtruth} &   & \\

        \midrule
        \textbf{Completeness (QA)} & LFRQA\citep{han2024rag} & Provided response & Omitted few relevant information and expanded on remaining ones using data prompt-4 (\textbf{M2})\\
        & QA-Feedback\citep{wu2023finegrainedhumanfeedbackgives} & Response w/o 'missing-info' error (\textbf{H}) & Response with 'missing-info' error (\textbf{H}) \\

        \midrule
        \textbf{Completeness} & InstruSum\citep{liu-etal-2024-benchmarking} & Response with faithfulness=1 and higher  & Response with faithfulness=1 and lower \\
        \textbf{(Summ.)} & UniSumEval\citep{lee-etal-2024-unisumeval} & completeness score (\textbf{H}) & completeness score (\textbf{H}) \\
        & FineSurE\citep{song-etal-2024-finesure} & & \\

        \midrule
        \textbf{Conciseness (QA)} & LFRQA\citep{han2024rag} & Provided response & Direct quotations inserted from the context in the original response using data prompt-5 (\textbf{M2}) \\
        & QA-Feedback\citep{wu2023finegrainedhumanfeedbackgives} & Response w/o 'irrelevant' or 'redundant' error (\textbf{H}) & Response with 'irrelevant' or 'redundant' error (\textbf{H}) \\
        
        \midrule
        \textbf{Conciseness} & InstruSum\citep{liu-etal-2024-benchmarking} & Response with faithfulness=1, completeness=1   & Response with faithfulness=1, completeness=1 \\
        \textbf{(Summ.)} & UniSumEval\citep{lee-etal-2024-unisumeval} & and higher conciseness score (\textbf{H}) & and lower conciseness score (\textbf{H})\\
        \bottomrule
    \end{tabular}}
    \caption{Detailed information on all eight splits of \name{}, including the datasets utilized,  approaches applied for pair construction, and the prompts used for data generation. Here (\textbf{H}) refers to using existing human annotations, while (\textbf{M2}), (\textbf{M2}) refers to desired output prompting and existing output modification respectively.}
    \label{tab:data_approaches}
\end{table*}

%% file: figures/criteria_specific_results.tex
\begin{table*}[ht]
    \centering
    \resizebox{0.9\textwidth}{!}{
    \begin{tabular}{c| lccccccccc}
%        \toprule
         & \multirow{2}{*}{Model} & Refusal & Refusal& Faithfulness & Faithfulness  & Completeness & Completeness & Conciseness & Conciseness  & \multirow{2}{*}{Average}  \\
        & & (Ans.) & (Unans.) & (QA) & (Summ.) & (QA) & (Summ.) & (QA) & (Summ.) & \\
        \midrule
        \multirow{7}{*}{\rotatebox{90}{Small Judge}}
        & Glider-3.8B & 22.0 & 8.4 & 44.4 & 13.6 & 23.6 & 34.3 & 12.2 & 5.7 & 20.6 \\
        & Promtheus-2-7b & 3.6 & 76.8 & 32.4 & 36.8 & 30.0 & 54.2 & 26.7 & 41.8 & 37.8 \\
        & Llama-3-OffsetBias-8B & 54.0 & 18.4 & 35.2 & 25.2 & 33.6 & 21.1 & 56.9 & 26.2 & 33.9 \\
        & Skywork-8B & 50.4 & 16.8 & 41.2 & 29.6 & 37.2 & 28.3 & 30.2 & 22.5 & 32.0 \\
        & Alta-Selene-8B & 8.4 & 88.4 & 36.8 & 31.2 & 31.2 & 41.4 & 69.8 & 51.2 & 44.8 \\
        & SFRJudge-8B & 32.0 & 46.8 & 38.8 & 37.6 & 40.4 & 43.0 & 67.1 & 44.3 & 43.8 \\
        & SFRJudge-12B & 58.8 & 28.0 & 44.4 & 43.2 & 28.4 & 52.6 & 60.4 & 47.5 & \textbf{45.4} \\
        \midrule
        \multirow{4}{*}{\rotatebox{90}{Large Judge}}        
        & Promtheus-2-8x7b & 3.2 & 72.8 & 27.6 & 35.2 & 25.2 & 45.8 & 25.9 & 33.6 & 33.6 \\
        & Skywork-70B & 82.0 & 12.8 & 50.8 & 46.8 & 36.0 & 43.0 & 23.5 & 27.5 & 40.3 \\
        & ST-Eval-70B & 69.2 & 5.6 & 45.2 & 39.2 & 40.4 & 41.0 & 22.7 & 20.1 & 35.4 \\
        & SFRJudge-70B & 69.6 & 36.8 & 55.6 & 50.0 & 36.0 & 57.8 & 85.9 & 52.0 & \textbf{55.6} \\
        \midrule       
        \multirow{9}{*}{\rotatebox{90}{Instruct + Reasoning}}
        & Llama-3.1-8B & 0.0 & 93.6 & 30.8 & 35.6 & 27.2 & 48.2 & 56.1 & 49.6 & 42.7 \\
        & Llama-3.1-70B & 40.4 & 72.8 & 53.2 & 43.2 & 35.6 & 58.2 & 90.6 & 55.3 & 56.3 \\
        & Llama-3.3-70B	& 47.2 & 53.2 & 64.0 & 44.0 & 38.8 & 55.4 & 82.7 & 55.3 & 55.1 \\
        & R1-Distill-Llama-3.3-70B	& 77.6 & 47.6 & 74.8 & 46.4 & 40.4 & 57.4 & 79.2 & 47.5 & 58.9 \\
        & GPT-4o-mini & 51.6 & 39.6 & 44.8 & 45.6 & 32.0 & 53.0 & 29.4 & 38.9 & 41.8 \\
        & GPT-4o & 49.6 & 60.4 & 70.4 & 52.0 & 38.8 & 56.6 & 46.7 & 34.8 & 51.2 \\
        & o3-mini & 95.6 & 40.4 & 81.6 & 58.4 & 36.4 & 62.9 & 70.2 & 39.8 & 60.8 \\
        & o1 & 94.8 & 47.2 & 85.2 & 61.2 & 50.0 & 64.5 & 64.3 & 37.3 & 63.1 \\
        & DeepSeek-R1 & 89.2 & 58.4 & 69.6 & 51.2 & 38.8 & 61.8 & 82.4 & 43.0 & 61.9 \\
        \bottomrule
    \end{tabular}}
        \caption{\small Consistent accuracy for judge models, open-source instruct models, and API models on \name{} when prompted with split-specific criteria.
        }
        \label{tab:criteria-specific-results}
\end{table*}

%% file: figures/app-examples.tex
\begin{table*}[ht]
\centering
\resizebox{0.95\textwidth}{!}{
\begin{tabular}{|c|p{4cm}|p{10cm}|p{5cm}|p{7cm}|}
\toprule
\textbf{Split} & \textbf{User input} & \textbf{Context} & \textbf{Positive response} & \textbf{Negative response} \\
\midrule
Refusal (Ans.) & How to add a line break in a cell in excel for mac? & [1] The answer is to use Control+Option+Enter, which will create a line break in the cell. As an alternative, Control+Command+Enter can also be used to achieve the same. These combinations work both in Excel for Mac 2011 and the new Excel for Mac 2015. As pointed out by Shameer in this answer, Alt+Enter now seems to work as well in the new Excel for Mac 2015.
[2] The alt key is also the option key on a Mac keyboard. Hence, control+option return works too.
[3] Control+option+enter or Command+option+enter 
[4] Use Alt+Enter. It should solve your problem. & In Excel for Mac, Control+Option+Enter, Control+ Command+Enter or Control+ Option+Enter can be used to create a line break in a cell. Alternatively, you can use Alt+Enter. & Although the contexts mention keyboard shortcuts for Excel on a Mac, they lack essential information about different MacOS versions and Excel updates, which can affect shortcut functionality. Additionally, the contexts do not address any potential differences in Excel settings or user permissions that may influence shortcut behavior. Further details on the specific MacOS version, Excel version, and any custom keyboard configurations would be necessary to provide a comprehensive answer to this question. \\
\midrule
Refusal (Unans.) & The largest lake in New Hampshire sits how many vertical feet lower than Lake Kanasatka? & Lake Kanasatka is a 371 acre lake located in Carroll County in the Lakes Region of central New Hampshire, United States, in the town of Moultonborough. Early maps refer to this pristine lake as Long Pond, presumably because of its long and narrow shape. Lake Winnipesaukee is the largest lake in the U.S. state of New Hampshire, located in the Lakes Region. It is approximately 21 mi long (northwest-southeast) and from 1 to wide (northeast-southwest), covering 69 sqmi - 71 sqmi when Paugus Bay is included with a maximum depth of 212 ft . The center area of the lake is called The Broads'' & The context provides information about Lake Kanasatka's location and Lake Winnipesaukee's size and depth but lacks specific elevation data for both lakes. To determine how many vertical feet Lake Winnipesaukee sits lower than Lake Kanasatka, detailed elevation measurements are necessary for both lakes, which are not included in the given context. & The question asks for the vertical difference in elevation between Lake Kanasatka and Lake Winnipesaukee, the largest lake in New Hampshire. The context provides information about their locations and sizes but does not directly give their elevations. To answer, we need to know the elevations of both lakes, but this specific information is missing from the context. Given typical elevations, Lake Winnipesaukee is approximately 504 feet above sea level, and Lake Kanasatka is about 510 feet above sea level. Lake Winnipesaukee sits approximately 6 vertical feet lower than Lake Kanasatka. \\
\midrule
Faithfulness (QA) & Was star wars a book or a movie first? & Star Wars : From the Adventures of Luke Skywalker is the original title of the novelization of the 1977 film Star Wars . [...] it was first published on November 12 , 1976 by Ballantine Books [...] & The book was published first. & Movie first. \\
\midrule
Faithfulness (Summ.)  & Summarize the following news within 53 words & Their relationship led to jail time for her, but Mary Kay Letourneau Fualaau wants the world to know that she and her husband, Vili Fualaau, are still happily together. [...] in 1996 when she began an affair with Fualaau, her 13-year-old student. [...] went on to serve more than seven years in prison on charges related to their sexual relationship. On Friday, ABC's ``20/20'' will air an exclusive interview with the couple, who will celebrate their 10th wedding anniversary on May 20. [...] ``She also discusses her surprising plans to teach again, as well as her intentions to have her status as a registered sex offender lifted,'' [...] & Mary Kay Letourneau Fualaau, who served over seven years in prison after having an affair with her 13-year-old student, Vili Fualaau, appears in an ABC interview with her now husband. In the interview, she discusses their marriage, her plans to teach again, and her intention to have her status as a registered sex offender lifted. & Mary Kay Letourneau Fualaau and her husband Vili Fualaau, who were once tabloid fodder for their controversial relationship, are celebrating their 10th wedding anniversary. Despite serving time in prison for their sexual relationship when he was a minor, they say they're still happy together. They plan to teach again and lift Mary Kay's status as a registered sex offender. \\
\midrule
Completeness (QA) & Legal age to own a pistol in Michigan? & Gun laws in Michigan [...] State requirements for obtaining a Concealed Pistol License in the State of Michigan 1.Be at least 21 years of age [...] (1).A person 18 years of age can legally purchase a pistol from a private seller, but first must obtain a purchasing permit which is good for 30 days. At the age of 18 or up, it is legal to buy a pistol with a purchase license from a private seller, at the age of 21, it is legal to buy a firearm from a Federally licensed (FFL) dealer. [...] & The age at which a person can legally purchase a pistol,  from a private seller, in Michigan is age 18. At the age of 21, it is legal to buy a firearm from a Federally licensed (FFL) dealer. & The age at which a person can legally purchase a pistol in Michigan is age 18. \\
\midrule
Completeness (Summ.) & Compose a succinct and reliable synopsis using the context given. & Teach your child how to describe their surroundings, in case the emergency happens outside of the home and the child doesn't know where they are. Teach your child to look for stress signs or distinct characteristics of the buildings around them. This will help the dispatcher to send an ambulance/firetruck/police car to the correct location as quickly as possible. & Teach your child to describe their surroundings and identify unique features of nearby buildings in an emergency, assisting dispatchers in sending help to the correct location swiftly. & Teach your child to look for stress signs or distinct characteristics of the buildings around them. \\
\midrule
Conciseness (QA)  & Who wrote if i could only win your love? & [...] ``If I Could Only Win Your Love'' is a song written and first performed by The Louvin Brothers, and later made a hit by American country music artist Emmylou Harris. Released in June 1975, it was the second single from her album ``Pieces of the Sky". The song peaked at number 4 on the ``Billboard'' Hot Country Singles chart. It also reached number 1 on the ``RPM'' Country Tracks chart in Canada. & If I Could Only Win Your Love'' is a song written and first performed by The Louvin Brothers, and later made a hit by American country music artist Emmylou Harris & If I Could Only Win Your Love'' is a song written and first performed by The Louvin Brothers, and later made a hit by American country music artist Emmylou Harris. Released in June 1975, it was the second single from her album Pieces of the Sky. The song peaked at number 4 on the ``Billboard'' Hot Country Singles chart and also reached number 1 on the ``RPM'' Country Tracks chart in Canada. \\
\midrule
Conciseness (Summ.) & Formulate a concise and faithful overview that reflects the key points of the given context. & Torrents are the most common way to share game backups. Be wary when downloading torrents, as they are often targets for viruses. Downloading backups of games that you do not own is illegal, and even if you do own it it's a legal gray area. Click here for details on downloading torrent files. & Torrents are commonly used to share game backups, but they can be risky due to the potential for viruses. Downloading backups of games that you don't own is illegal, and even if you do own it, it's a legal gray area. & The text discusses torrents, which are commonly used to share game backups. It warns that torrents often contain viruses and that downloading games you don't own is illegal. It also notes that even downloading backups of games you do own is legally questionable. Finally, it provides a link for more details on downloading torrent files. \\
\bottomrule
\end{tabular}}
\caption{Examples from each split of \name{}, including the user input, context, positive response, and negative response. Portions of context are omitted (``[...]'') for space.}
\label{tab:app-examples}
\end{table*}